\pdfoutput=1
\documentclass[10pt,twocolumn,letterpaper]{article}

\usepackage[pagenumbers]{cvpr}
\usepackage{multirow}
\usepackage{wrapfig}
\usepackage{placeins}

\definecolor{cvprblue}{rgb}{0.21,0.49,0.74}
\usepackage[breaklinks,colorlinks,allcolors=cvprblue]{hyperref}
\hypersetup{
  pdftitle={GraspHOI: Full-Body 3D Human-Object Reconstruction with Articulated Hands from a Single In-the-Wild Image},
  pdfauthor={Semin Kim, Haechan Shin, Jongyoo Kim}
}

\def\confName{CVPR}
\def\confYear{2027}

\title{GraspHOI: Full-Body 3D Human-Object Reconstruction with Articulated Hands from a Single In-the-Wild Image}
\author{Semin Kim \qquad Haechan Shin \qquad Jongyoo Kim\\
Yonsei University\\
{\tt\small \{hohosemin,haechanshin,jy.kim\}@yonsei.ac.kr}
}

\makeatletter
\apptocmd{\@maketitle}{%
  \begin{minipage}{\textwidth}
    \centering
    \includegraphics[width=\textwidth,keepaspectratio]{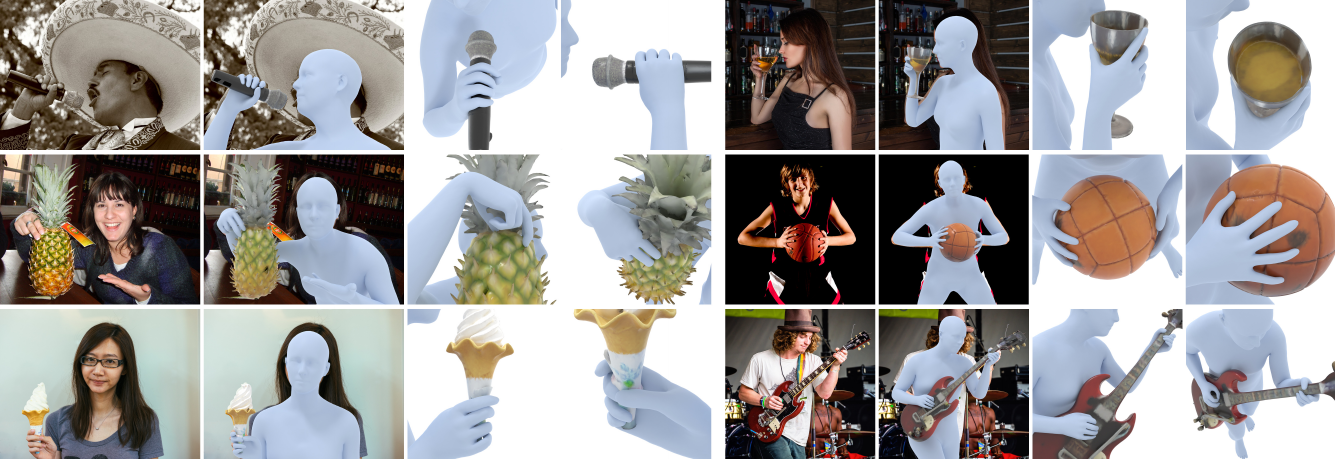}
    \captionsetup{hypcap=false}
    \captionof{figure}{\textbf{GraspHOI.} Given a monocular RGB image, our
    method reconstructs human-object interactions with articulated hands and
    realistic poses across diverse in-the-wild images.}
    \label{fig:teaser}
  \end{minipage}
  \vspace{6pt}%
}{}{\PackageError{GraspHOI}{Could not attach Figure 1 to the title block}{}}
\makeatother

\begin{document}
\maketitle

\begin{abstract}
Existing monocular full-body 3D human-object interaction (HOI) methods do not combine category-agnostic object reconstruction with finger-level contact. Their hands may float off or sink into objects, breaking realism in AR/VR environments. We present \mbox{\textbf{GraspHOI}}, the first category-agnostic framework that reconstructs a full-body 3D HOI from a single image and jointly refines the arms and fingers so that the hands meet the reconstructed object surface. GraspHOI reconstructs the body, hands, and object separately with foundation models and places them in a shared metric camera frame through depth-based registration and silhouette alignment. Occlusion-aware contact correspondences seat the object against the hand, and contact-aware optimization refines shoulder, elbow, wrist, and finger rotations while anchoring the joints to the image. We find that with fingers capped near the image estimate, refining them alone is insufficient; freeing the arm removes most of the remaining hand-object gap, since in a full body the arm chain places the hand. In experiments on three datasets, GraspHOI yields more plausible hand-object contact than four baselines and better relative human-object placement than HDM and TeHOR.
\end{abstract}
\vspace{-8pt}
\section{Introduction}
\label{sec:intro}

Full-body human-object interaction (HOI) reconstructions can serve as editable assets for animation, game cutscenes, and virtual reality, and as keyframes for contact-plausible HOI video generation. Useful reconstructions must cover in-the-wild objects without category constraints and recover articulated hands within the full-body interaction. A plausible body-object configuration alone is insufficient as category-bound geometry limits the interactions represented, while fingers that penetrate an object or float from its surface break the illusion and yield poor keyframes for downstream grasp motion.

Existing full-body methods do not combine category-agnostic object reconstruction with hand articulation optimization. Template-based approaches rely on predefined object meshes~\cite{zhang2020phosa, xie2022chore, xie2023vistracker}. Template-free methods~\cite{xie2023template_free, xie2024intertrack} instead reconstruct human-object point clouds, freeing the model of template objects, but their learned category priors remain and neither explicitly predicts finger articulation. Retrieval-based methods~\cite{Cseke2025PICO,dwivedi2025interactvlm} broaden category coverage but remain database dependent, and do not directly reconstruct the observed object geometry. CARI4D~\cite{xie2025cari4d} achieves category-agnostic full-body 4D reconstruction from video, but represents contact with two hand-joint locations rather than optimized finger articulation. Hand-object interaction works~\cite{Liu_2025_CVPR,fan2024hold} optimize fingers directly against geometry, but do not take arms into consideration, a critical component in human-object interaction.

\textbf{GraspHOI} is the \textbf{first category-agnostic framework that reconstructs a full-body 3D HOI from a single image while explicitly optimizing hand articulation against the reconstructed object}. Here, category-agnostic denotes freedom from predefined object meshes, category priors, or contact retrieval. The object, body, and hands are reconstructed independently, using foundation models. Although the individual meshes may each project plausibly into the image, they can disagree in metric scale and camera-space translation. We therefore place them in a shared metric camera frame and use explicit surface correspondences to refine their relative 3D configuration. Our method refines both arms and hands, favoring plausible contact over per-joint accuracy. Temporal motion and body-object interactions such as sitting or leaning are beyond our present scope.

\begin{figure*}[t]
  \centering
  \includegraphics[width=\textwidth]{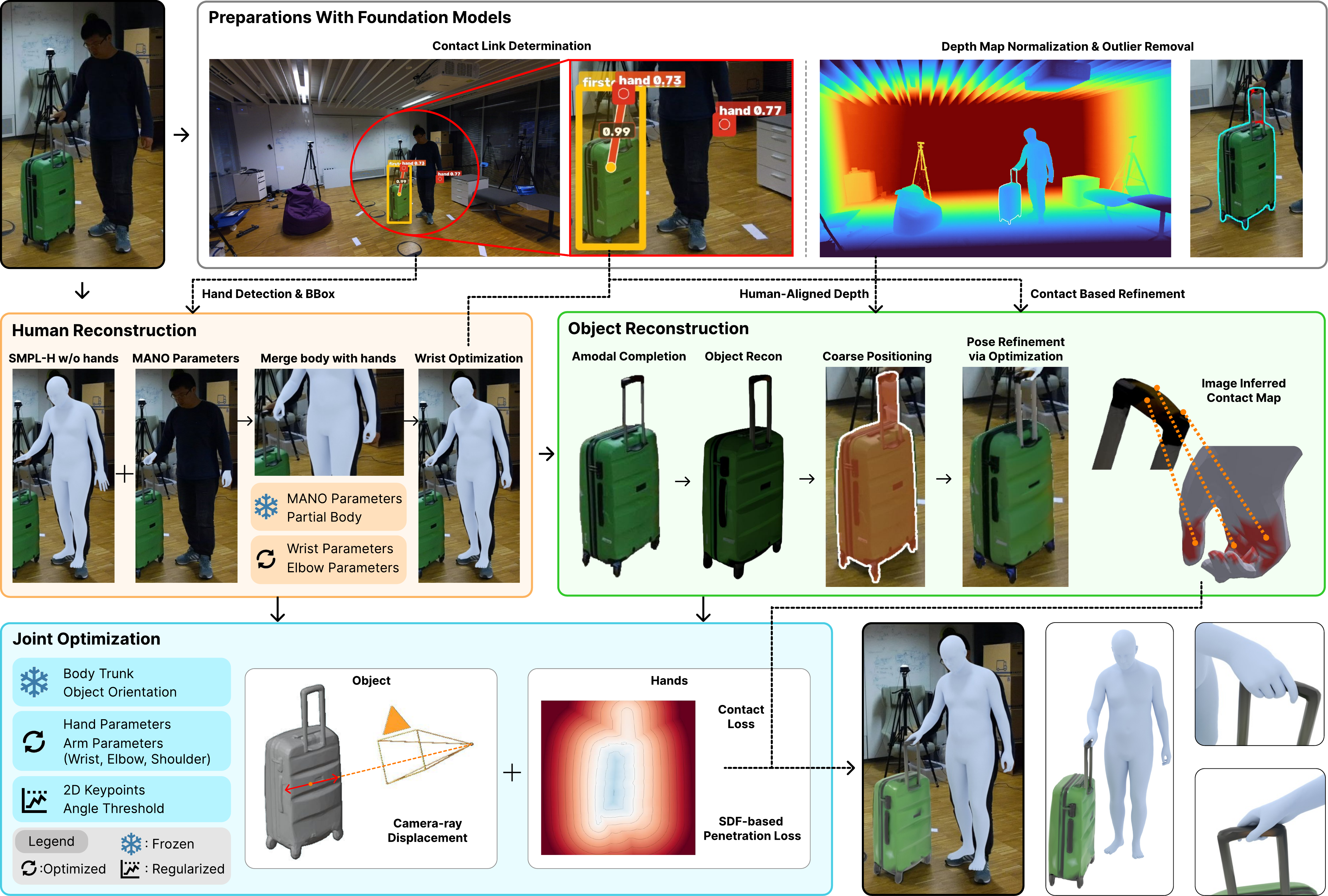}
  \caption{GraspHOI reconstructs the human, hands, and object separately, places them in a shared metric camera frame, and refines their relative configuration through contact seating and joint optimization. The input image is from InterCap~\cite{huang2024intercap}.}
  \label{fig:big}
\end{figure*}

In detail (Fig.~\ref{fig:big}), SAM 3D Body~\cite{Yang_2026_CVPR} reconstructs the body in the Momentum Human Rig (MHR)~\cite{ferguson2025mhr}. We fit its mesh to SMPL-H and insert articulated hands recovered by
WiLoR~\cite{potamias2024wilor}. SAM 3D Objects~\cite{sam3dteam2025sam3d3dfyimages} reconstructs the object from an amodally completed image (Sec.~\ref{sec:method_pre}). Its predicted layout, with metric RGB-D registration as a fallback, initializes the object pose, which we then fit to the silhouette and scale to metric depth (Sec.~\ref{sec:method_obj}). Occlusion-aware correspondences then seat the object against the hand (Sec.~\ref{sec:method_contact}), after which contact-aware optimization refines arm and finger articulation together with the object's depth along camera rays (Sec.~\ref{sec:method_jointopt}).

We evaluate GraspHOI against four recent baselines, HDM~\cite{xie2023template_free}, TeHOR~\cite{Nam_2026_CVPR}, EasyHOI~\cite{Liu_2025_CVPR}, and PICO~\cite{Cseke2025PICO}, across three
datasets: ARCTIC~\cite{Fan_2023_CVPR}, our ProciGen-GRAB (GRAB~\cite{Taheri2020GRAB} interactions rendered with the ProciGen~\cite{xie2023template_free} pipeline), and PICO-db~\cite{Cseke2025PICO}. The evaluation covers joint HOI recovery, hand pose, and contact plausibility, and ablations study the main stages.

Our key contributions are as follows:
\begin{itemize}
    \item The first category-agnostic single-image framework that reconstructs a full-body 3D HOI and jointly refines the arms and fingers against the reconstructed object surface.
    \item An empirical analysis showing that, with finger articulation capped near the image estimate, refining the arm chain closes most of the remaining hand-object gap in full-body reconstruction.
\end{itemize}

\noindent Our code will be publicly released.

\section{Related Work}
\label{sec:related}

\textbf{Human Body Reconstruction.}
Representative SMPL-family reconstruction methods include~\cite{Bogo_2016_ECCV, Kanazawa_2018_CVPR, Kolotouros_2019_ICCV, Zhang_2021_ICCV, Kocabas_2021_ICCV, Li_2022_ECCV, patel2024camerahmr, sarandi2024nlf}. The underlying SMPL model~\cite{loper2015smpl} lacks articulated hands, limiting fine-grained interaction modeling. SMPL-H/SMPL-X methods~\cite{pavlakos2019smplx, cai2023smplerx, yin2025smplestx, zhang2023pymafx, Feng_2021_PIXIE, Rong_2021_FrankMocap} add articulated hands, while the more recent SAM 3D Body~\cite{Yang_2026_CVPR} instead predicts MHR~\cite{ferguson2025mhr}. We use SAM 3D Body for full-body initialization and refine hand articulation separately.

\textbf{3D Object Reconstruction.}
Single-image object reconstruction has used generated multiview observations with implicit reconstruction~\cite{liu2023one2345, liu2023one2345plusplus}, SDS optimization~\cite{poole2022dreamfusion, Lin_2023_CVPR_Magic3D}, feed-forward models~\cite{hong2023lrm, tochilkin2024triposr, Xu_2024_InstantMesh}, and diffusion-guided explicit 3D generation~\cite{Tang_2023_DreamGaussian, Long_2024_Wonder3D}. Large-scale Hunyuan3D models~\cite{yang2024hunyuan3d,hunyuan3d2_2025,hunyuan3d21_2025,hunyuan3d2025hunyuan3d,lai2025hunyuan3d} improve fidelity across unseen categories, while TRELLIS~\cite{xiang2025structured, xiang2025native} provides scalable structured latents. SAM 3D Objects~\cite{sam3dteam2025sam3d3dfyimages}, which is used in our work, reconstructs complete geometry and texture under occlusion while jointly predicting scene layout.

\textbf{Human-Object Interaction.}
Template-based HOI methods~\cite{zhang2020phosa, xie2022chore, xie2023vistracker, nam2024contho} require predefined object meshes, limiting generalization beyond known geometries. PHOSA~\cite{zhang2020phosa} already constrains object scale with the interaction, which our seating extends to reconstructed meshes. HOI-TG~\cite{Wang2025HOI-TG} regresses the interacting body and object end-to-end with a graph transformer, and ScoreHOI~\cite{li2025scorehoi} applies score-guided diffusion with contact and penetration guidance, but both remain template-bound. Template-free HDM~\cite{xie2023template_free} adapts PC$^2$~\cite{melaskyriazi2023projection} for single-image point-cloud reconstruction, and InterTrack~\cite{xie2024intertrack} extends this representation across video with temporal consistency. Both remove templates, but their learned category priors remain, and neither explicitly predicts finger articulation. CARI4D~\cite{xie2025cari4d} is the first category-agnostic full-body 4D HOI method for monocular video. However, it assumes substantial initial object visibility.

Alternative paradigms use vision–language priors for interaction semantics~\cite{dwivedi2025interactvlm, zhang2025openhoi, zhang2025interactanything}. TeHOR~\cite{Nam_2026_CVPR} aligns textured human-object reconstructions with textual interaction descriptions. Others use human motion or physics for object pose~\cite{petrov2023object, zhang2025force, zhao2026real2sim}, and diffusion priors for globally plausible joint synthesis~\cite{petrov2025tridi, ron2025hoidini, zhang2024hoidiffusion}. Wen et al.~\cite{wen2025open3dhoi} fit open-vocabulary HOI with part-level contact, and grasp synthesis~\cite{taheri2022goal,wu2022saga,tendulkar2023flex} generates rather than reconstructs whole-body grasps. Their focus is global interaction semantics or generation rather than refining finger articulation against reconstructed object contact.

Contact predictors~\cite{tripathi2023deco, bierling2025decodino, yang2024lemon} provide contact maps over broader object sets. PICO~\cite{Cseke2025PICO} retrieves an object mesh and the closest contact correspondences from PICO-db. This broadens category coverage, but generalization depends on compatible retrievals from finite databases. GraspHOI instead reconstructs object geometry and establishes contact geometrically, enabling detailed grasp optimization without predefined meshes or retrieved interaction examples. Hand-object methods~\cite{hasson2019obman,grady2021contactopt,jiang2021grasptta,cao2021rho,ye2022ihoi,zhou2022toch,ye2024ghop} model only the hand and object, placing the hand freely rather than through a full-body arm chain.

\section{Method}
\label{sec:method}

We introduce GraspHOI, a hand-articulation-aware, category-agnostic full-body HOI reconstruction method. Besides the image $I$, it takes an object mask $S_{\mathrm{vis}}$ (\eg, from SAM 3~\cite{carion2025sam}) and the object's name as input. Hand detections and interaction
links are inferred from $I$. Foundation models recover the human, hands, and
object. Each model outputs meshes and their poses in respective frames that differ in scale or viewpoint.

The subsequent stages solve cross-component scale, pose, and surface contact in
a shared metric camera frame for body and object. Because the losses act directly on the reconstructed surfaces, stages can correct contact and penetration errors after initialization.

\subsection{Human and Object Initialization}
\label{sec:method_pre}

\textbf{Human reconstruction.}
We estimate camera intrinsics $K$ with HumanFoV, the field-of-view estimator
released with CameraHMR~\cite{patel2024camerahmr}, and reconstruct the body
with SAM 3D Body~\cite{Yang_2026_CVPR}. All foundation models that take intrinsics share them. We use its body-only, hand-frozen
inference setting to predict a camera-space MHR~\cite{ferguson2025mhr} mesh.
Following the official MHR-to-SMPL-X converter, we adapt its barycentric topology transfer to a
neutral SMPL-H target. This recovers SMPL-H global orientation, body pose
$\boldsymbol\theta_b$, shape $\boldsymbol\beta$, and translation, while finger
pose remains neutral. HOI-DETR~\cite{darkhalil2026improving} then detects hand boxes and
hand-object interaction links. We associate each detected box with the
projected left or right SMPL-H hand and apply
WiLoR~\cite{potamias2024wilor} to recover MANO~\cite{Romero2017MANO}
finger poses and 2D hand keypoints. We insert these hand estimates into the fitted SMPL-H body via inverse kinematics. No vertices are copied between MANO and SMPL-H and only pose parameters are transferred.

\textbf{Object reconstruction and amodal conditioning.}
We first estimate a dense metric depth map $D$ with Pixel-Perfect Depth
(PPD)~\cite{xu2025pixel}, using MoGe-2~\cite{wang2025moge2} as its geometry prior. PPD and SAM 3D Body estimate metric depth independently, so we align them once before placement, at the hands, where the object is placed. Over the camera-facing
vertices $v$ of the detected hands whose projected pixels $x_v$ lie outside
$S_{\mathrm{vis}}$, so that mostly hand surface is sampled, we set
$\gamma=\operatorname{median}_{v}[v_z/D(x_v)]$ and replace $D\leftarrow\gamma D$.
When too few hand vertices qualify, we use all human vertices instead.
This hand-aligned depth map is shared by all later placement and contact stages.

We reconstruct the object with SAM 3D Objects~\cite{sam3dteam2025sam3d3dfyimages},
which predicts a textured mesh and a layout token for pose and position. The layout is not accurate enough to place the object directly (Sec.~\ref{sec:method_obj}), so we optimize the placement
against the object silhouette. The visible mask $S_{\mathrm{vis}}$ is a poor
target due to occluders, especially the hands. We therefore complete the object amodally. The
depth map identifies pixels lying in front of the visible object within a local
object region. Only this inferred occluder is inpainted, with Stable Diffusion~\cite{rombach2021highresolution}, while visible object
pixels are preserved. We obtain an amodal cutout $I_a$ and silhouette $S_a$. SAM 3D Objects receives $S_a$ and the image with $I_a$ pasted into $S_a\setminus S_{\mathrm{vis}}$, and $S_a$ becomes the target of every placement step downstream. The reconstructed mesh is $\mathcal M_o^0=(V_o^0,F_o)$, where $V_o^0$ and $F_o$ are its vertices and faces and the superscript $0$ denotes the initial state.

\subsection{Metric Object Placement}
\label{sec:method_obj}

SAM 3D Objects predicts its layout based on the camera of its internal
depth estimator. The predicted shape and orientation are largely reliable, but the position is not. We therefore treat the layout as an initialization, with metric RGB-D registration as a
fallback, and optimize the placement from either one.

\textbf{Translation re-fit.}
Let $\widehat S$ be the silhouette of $V_o$ rendered under $K$, and $|S|$ the
area of a silhouette $S$. We translate $V_o$ so that the centroid of $\widehat S$
moves onto that of $S_a$ and its depth is multiplied by
$\rho=(|\widehat S|/|S_a|)^{1/2}$. Under weak perspective, \ie, when the object's
depth range is small relative to its distance, this step matches both the
centroid and the area exactly; under full perspective, repeating it shrinks the
residual error geometrically. We repeat the step as long as it raises the IoU with $S_a$.

\textbf{Gate and fallback.}
We gate the re-fitted layout at $\operatorname{IoU}(\widehat S,S_a)\ge0.7$, a common threshold for a confident overlap. For layouts below the threshold, the same mesh is also registered with
FoundationPose~\cite{wen2024foundationpose} on an as-if-unoccluded RGB-D
observation built from the amodally completed object cutout $I_a$, and its pose is re-fitted as above. Both placements then pass through the metric scale and silhouette refinement steps, and we keep the one with the higher final IoU with $S_a$. The threshold thus trades runtime for accuracy.

\textbf{Coarse metric scale.}
We scale $V_o$ by a scalar $N_d$ along camera rays, so scale and translation change together but the projection does not. $N_d$ is the median ratio of $D$ to the
rendered depth of $V_o$ over the visible object pixels.

\textbf{Silhouette refinement.}
Finally, we refine the rotation, translation, and scale of the placed mesh.
With $\bar V_o$ the placed mesh centered at its centroid $\bar c$, we optimize
$V_o=sR(q)\bar V_o+t$ over the axis-angle rotation $q\in\mathbb R^3$, the
translation $t\in\mathbb R^3$, and the scale $s>0$, starting from $q=0$,
$t=\bar c$, and a rendered-to-target box correction of $s$. Here $\widehat S$
is rendered in a mask-centered crop, with $K$ transformed to the crop
coordinates. Let $b_c(\cdot)$ and $b_s(\cdot)$ denote the center and size of a
2D bounding box. Scale and position are constrained by soft barriers,
\begin{equation}
\begin{split}
\mathcal L_{\mathrm{sc}}&=
\left[\|b_s(\widehat S)-b_s(S_a)\|_2^2-\epsilon_s^2\right]_+,\\
\mathcal L_{\mathrm{pos}}&=
\left[\|b_c(\widehat S)-b_c(S_a)\|_2^2-\epsilon_p^2\right]_+,
\end{split}
\label{eq:object_barriers}
\end{equation}
where $[x]_+=\max(x,0)$, and $\epsilon_s$ and $\epsilon_p$ are respectively
$5\%$ and $3\%$ of the target box diagonal. $\epsilon_p$ is tight, anchoring the box center until the position is close, and $\epsilon_s$ is looser, since the reconstructed shape may not match the image exactly. These terms prevent drift without
competing with fine boundary alignment.

The symmetric silhouette discrepancy is
\begin{equation}
\mathcal L_{\mathrm{sil}}=
\left[100\left(
\frac{\sum S_a(1-\widehat S)}{\sum S_a}+
\frac{\sum \widehat S(1-S_a)}{\sum \widehat S}
\right)\right]^2.
\label{eq:silhouette_loss}
\end{equation}
We additionally use a symmetric contour objective, which is intended for thin structures. Let
$C_a$ be the target contour-pixel set, $C_m(x)$ the predicted soft-contour
intensity, and $D_a(x)$ the pixel distance to $C_a$. Then
\begin{equation}
\mathcal L_{\mathrm{cnt}}
=\frac{\langle C_m,D_a^2\rangle}{\|C_m\|_1}
+\frac{\eta}{|C_a|}\sum_{x\in C_a}
\bigl(1-\widetilde C_m(x)\bigr)^2.
\label{eq:contour_loss}
\end{equation}
Here $\widetilde C_m=C_m/\max_x C_m(x)$ and $\eta=400$. The first term pulls
predicted boundaries toward the observation, while the reverse coverage term
prevents a smaller silhouette from matching only a subset of the target
boundary. The object objective is
\begin{equation}
\mathcal L_{\mathrm{obj}}=
10(\mathcal L_{\mathrm{sc}}+\mathcal L_{\mathrm{pos}})+
\mathcal L_{\mathrm{sil}}+w_c\mathcal L_{\mathrm{cnt}},
\label{eq:object_total_loss}
\end{equation}
with $w_c$ increased linearly from $0.3$ to $1$, so the mesh first targets the broad silhouette and then its fine details.

\subsection{Occlusion-Aware Contact Seating}
\label{sec:method_contact}

Object placement aligns the image evidence but does not determine which hand
surface touches which part of the object. We treat each HOI-DETR detection as a binary contact indicator for the associated hand. The target object is
identified by matching the linked object detection to silhouette $S_a$, and a contact is
discarded if the associated hand box has negligible overlap with the amodal
object region, preventing a hand from being linked to another nearby object. The hands in contact form the set $\mathcal H_c$ of contacting hands.

For each contacting hand $g$ in the set $\mathcal H_c$, we cast camera rays
through the overlap of its projection and $S_a$. Intersections with the SMPL-H
hand mesh are filtered by the palm normal, removing dorsal candidates and
retaining palmar-facing vertices $h_i$. Each retained vertex
is paired with an object vertex $o_j$ on the same image ray. The visible object
mask resolves the near-far ambiguity. Contact lies on the rear surface where
the object is visible and on the front surface where the hand occludes it. The resulting index pairs $(i,j)$ form the correspondence set $\mathcal C_g$ of hand $g$, and $\mathcal C$ collects the pairs of all contacting hands.

The object is then seated by a positive scalar $N$ about the camera origin.
For perspective projection $\Pi_K$ under $K$, we have
$\Pi_K(No)=\Pi_K(o)$, so this operation changes metric size and depth while keeping the object projection identical throughout. We solve
\begin{equation}
\begin{aligned}
N^*=\arg\min_N\quad &
\frac{1}{|\mathcal C|}\sum_{(i,j)\in\mathcal C}
\bigl(N z(o_j)-z(h_i)\bigr)^2 \\
&+\lambda_{\mathrm{seat}}\mathcal L_{\mathrm{pen}}^{\mathrm{seat}}(N),
\end{aligned}
\label{eq:depth_seating}
\end{equation}
over a bounded one-dimensional grid, where $z(\cdot)$ extracts camera-space
depth. The first pass uses $\lambda_{\mathrm{seat}}=0$. If maximum penetration
exceeds $2$\,mm, a second pass weights the mean squared hand-in-object depth
$\mathcal L_{\mathrm{pen}}^{\mathrm{seat}}$ as detailed in the supplement.
These fixed one-to-one pairs are passed to joint refinement. A nearest-surface
fallback is used only when fewer than three valid pairs are available.
We update $V_o\leftarrow N^* V_o$ and define the seated mesh
$\mathcal M_o=(V_o,F_o)$, so $o_j$ denotes a seated object vertex below.

\subsection{Contact-Aware Joint Refinement}
\label{sec:method_jointopt}

The final stage refines the shoulder, elbow, wrist, and finger rotations
for contacting hands. The body trunk and object orientation remain fixed, while the object's scale and translation change together through one scalar $s_o$ ($V_o'=s_o V_o$), which, as in seating, keeps its projection unchanged.

\textbf{Image anchoring.}
For each contacting hand $g\in\mathcal H_c$, let $u_g$ be the image projections
of its SMPL-H finger joints and $a_g$ those of its shoulder, elbow, and wrist.
A superscript $0$ marks values before refinement. The loss
\begin{equation}
\mathcal L_{\mathrm{kp}}=
\sum_{g\in\mathcal H_c}\left[
\operatorname{MSE}(u_g,\widehat p_g)+
\operatorname{MSE}(a_g,a_g^0)\right]
\label{eq:keypoint_loss}
\end{equation}
pulls the fingers onto the WiLoR keypoints $\widehat p_g$ and holds the arm
where it started in the image, preventing drift and enforcing faithfulness to the input image.

\textbf{Paired contact and penetration.}
The seated correspondences remain fixed during optimization. We define the
per-hand loss $\ell_g$ and average it across contacting hands:
\begin{equation}
\begin{aligned}
\ell_g&=\frac{1}{|\mathcal C_g|}\sum_{(i,j)\in\mathcal C_g}
\|h_i-s_o\,o_j\|_2^2,\\
\mathcal L_{\mathrm{cont}}&=
\frac{1}{|\mathcal H_c|}\sum_{g\in\mathcal H_c}\ell_g.
\end{aligned}
\label{eq:contact_loss}
\end{equation}
We use an object SDF that is positive outside and negative inside, and allow a
small compliance margin $\epsilon_{\mathrm{pen}}$ before penalizing penetration:
\begin{equation}
\mathcal L_{\mathrm{pen}}=10\mathcal L_{\mathrm{tun}}+
\sum_{v\in V_{\mathrm{hand}}}
\left[-\operatorname{SDF}_{s_o\mathcal M_o}(v)-
\epsilon_{\mathrm{pen}}\right]_+^2.
\label{eq:penetration_loss}
\end{equation}
Here $V_{\mathrm{hand}}$ contains the vertices of all contacting hands.
We set $\epsilon_{\mathrm{pen}}=2$\,mm to model soft-tissue compliance. Thin objects such as cup walls and book
pages leave the SDF too little interior to push a finger out, so
$\mathcal L_{\mathrm{tun}}$ pulls the joint just past the first
surface crossing back to the wrist side. A 2D trust region $\mathcal L_{\mathrm{tr}}$ keeps the hand
and arm joints near their initial projections, so that these terms cannot drag
the hand off its image evidence. Supplementary S2.7 details both.

\textbf{Pose regularization.}
Over the refined joints $k$, we use
\begin{equation}
\mathcal L_{\mathrm{reg}}=\sum_k\Bigl(\omega_k\,d_k^2+\kappa_k\,[d_k-c_k]_+^{p_k}\Bigr),
\label{eq:regularization_loss}
\end{equation}
where $d_k$ is the geodesic angle between the current and initial rotations of
joint $k$. The shoulder, elbow, and wrist rotations are softly capped at
$c_k=15^\circ$, $20^\circ$, and $30^\circ$ from their initial values,
respectively, with $p_k=2$. These values were set to allow large movements if necessary, while preventing drift. Finger joints use a steeper $p_k=4$ and a small
shared $c_k=3^\circ$, since $\mathcal L_{\mathrm{kp}}$ cannot resist the contact pull
along camera rays. The weights $\omega_k$ and $\kappa_k$ take one value for arm
joints and one for finger joints.
The final refinement objective is
\begin{equation}
\begin{aligned}
\mathcal L_{\mathrm{joint}}=
{}&\lambda_{\mathrm{kp}}\mathcal L_{\mathrm{kp}}+
\lambda_{\mathrm{cont}}\mathcal L_{\mathrm{cont}}\\
&+\lambda_{\mathrm{pen}}(\mathcal L_{\mathrm{pen}}+30\mathcal L_{\mathrm{tr}})+
\lambda_{\mathrm{reg}}\mathcal L_{\mathrm{reg}}.
\end{aligned}
\label{eq:joint_total_loss}
\end{equation}
The four $\lambda$ coefficients are stage-dependent scalar weights. 
For two-hand interactions, a coarse stage first optimizes both arm chains,
keeping the fingers fixed, to resolve front-back arm ambiguity. The standard refinement uses
$(\lambda_{\mathrm{kp}},\lambda_{\mathrm{cont}},\lambda_{\mathrm{pen}},\lambda_{\mathrm{reg}})=(0.01,3\times10^4,1.5\times10^5,100)$
for its 500 iterations, while the coarse stage uses $(1,10^4,\allowbreak{5\times10^5},100)$.
Detailed design choices are reported in the supplementary material.

\section{Experiments}
\label{sec:experiments}

\begin{table*}[!t]
\centering
\scriptsize
\setlength{\tabcolsep}{3.2pt}
\resizebox{\textwidth}{!}{%
\begin{tabular}{llcccccccc}
\toprule
& & \multicolumn{2}{c}{Human} & \multicolumn{1}{c}{Object}
& \multicolumn{1}{c}{H+O} & \multicolumn{4}{c}{Hands} \\
\cmidrule(lr){3-4}\cmidrule(lr){5-5}\cmidrule(lr){6-6}\cmidrule(lr){7-10}
Dataset & Method & sCD $\downarrow$ & MPJPE $\downarrow$
& sCD $\downarrow$ & sCD $\downarrow$
& MPJPE $\downarrow$ & PA-MPJPE $\downarrow$ & Offset (FL / PD) $\downarrow$ & PJO $\downarrow$ \\
\midrule
\multirow{3}{*}{ARCTIC}
& HDM$^\dagger$ & 72.73 & 140.13 & 179.15 & 94.58
& 165.46 & 9.49 & 56.62 (53.14 / 3.49) & 82.58 \\
& TeHOR & 66.13 & 130.10 & 747.36 & 116.08
& 132.53 & 9.01 & 608.69 (608.67 / 0.02) & 662.75 \\
& GraspHOI (Ours) & \textbf{26.66} & \textbf{56.24} & \textbf{43.08} & \textbf{30.82}
& \textbf{79.68} & \textbf{6.50} & \textbf{2.14} (0.34 / 1.80) & \textbf{13.30} \\
\midrule
\multirow{3}{*}{PGG}
& HDM$^\dagger$ & 29.93 & 61.17 & 173.64 & 73.39
& 163.64 & 10.25 & 64.16 (61.62 / 2.55) & 88.73 \\
& TeHOR & 40.72 & 85.19 & 769.04 & 124.66
& 136.65 & 8.83 & 615.82 (615.80 / 0.02) & 662.00 \\
& GraspHOI (Ours) & \textbf{29.03} & \textbf{59.72} & \textbf{126.59} & \textbf{61.34}
& \textbf{90.81} & \textbf{7.01} & \textbf{6.16} (4.43 / 1.73) & \textbf{17.22} \\
\bottomrule
\end{tabular}}
\caption{HOI baseline comparison on ARCTIC and PGG frames. sCD and MPJPE use the shared pelvis alignment. $^\dagger$ denotes human-scale alignment for up-to-scale HDM. TeHOR's public code omits the ICP step of its paper setup. All metrics are in mm.}
\label{tab:hoi_baselines}
\vspace{-2.5mm}
\end{table*}

\subsection{Evaluation Protocol}
\label{sec:protocol}

\textbf{Datasets.}
ARCTIC~\cite{Fan_2023_CVPR} provides calibrated multi-view RGB with SMPL-X body,
hand, and articulated-object meshes. We subsample
its interaction sequences at 0.5\,fps, yielding approximately 5K monocular
frame-views.

GRAB~\cite{Taheri2020GRAB} provides metric SMPL-X bodies, articulated hands,
and object meshes but no captured RGB. We therefore construct
\textbf{ProciGen-GRAB (PGG)} by feeding GRAB interactions through the
ProciGen~\cite{xie2023template_free} rendering pipeline in place of its original
source interactions. GRAB supplies the posed body and
hands together with the object trajectory, preserving their relative 3D
configuration, while ProciGen supplies clothed appearance and calibrated
monocular rendering. Subsampling at 0.5\,fps gives approximately 4K PGG
frame-views. PICO-db~\cite{Cseke2025PICO} is used for the PICO comparison as well as qualitative examples.

\textbf{Metrics.}
We sample 10K surface points from each mesh, and report shape Chamfer distance (sCD) for the human, the object, and their combined samples (H+O). Evaluation is performed with pelvis alignment, where one translation is shared by the human and
object, preserving their relative placement. Only PA-MPJPE uses rotation (Procrustes) alignment. Since HDM predicts point clouds, we fit SMPL-H to its human points via InterTrack's correspondence autoencoder~\cite{xie2024intertrack} and evaluate its human metrics on this fit. Because HDM is up-to-scale, we match its SMPL body-bone length to ground truth. In Table~\ref{tab:hoi_baselines}, we evaluate shape and 3D placement with pelvis-relative sCD and body MPJPE over the 22 SMPL body joints. Hand MPJPE uses the wrist and 15 finger joints of each annotated contacting hand in the same pelvis-aligned frame, either without further alignment or with per-hand Procrustes alignment (PA-MPJPE). Among 3D hand-pose errors, Table~\ref{tab:optimization_baselines} reports only PA-MPJPE, because EasyHOI reconstructs a single hand in its own camera frame; some metrics are omitted on PICO-db, which provides no posed 3D hand joints. Float (FL) is the gap between a non-penetrating hand and the object, and penetration depth (PD) is the mean depth of a penetrating hand's vertices inside the object. Both are averaged over hands in selected frames that have contact. The sum of FL and PD is Offset, measuring contact plausibility against each method's own object, and Contact (\%) is the percentage of hands whose Offset is at most 3\,mm. Per-Joint Offset (PJO) is the absolute difference between predicted and ground-truth signed distances to the object, averaged over 21 hand keypoints. For optimization-based methods, we additionally provide 2D MPJPE, as they optimize directly against image evidence.

\begin{figure*}[!t]
  \centering
  \includegraphics[width=\textwidth]{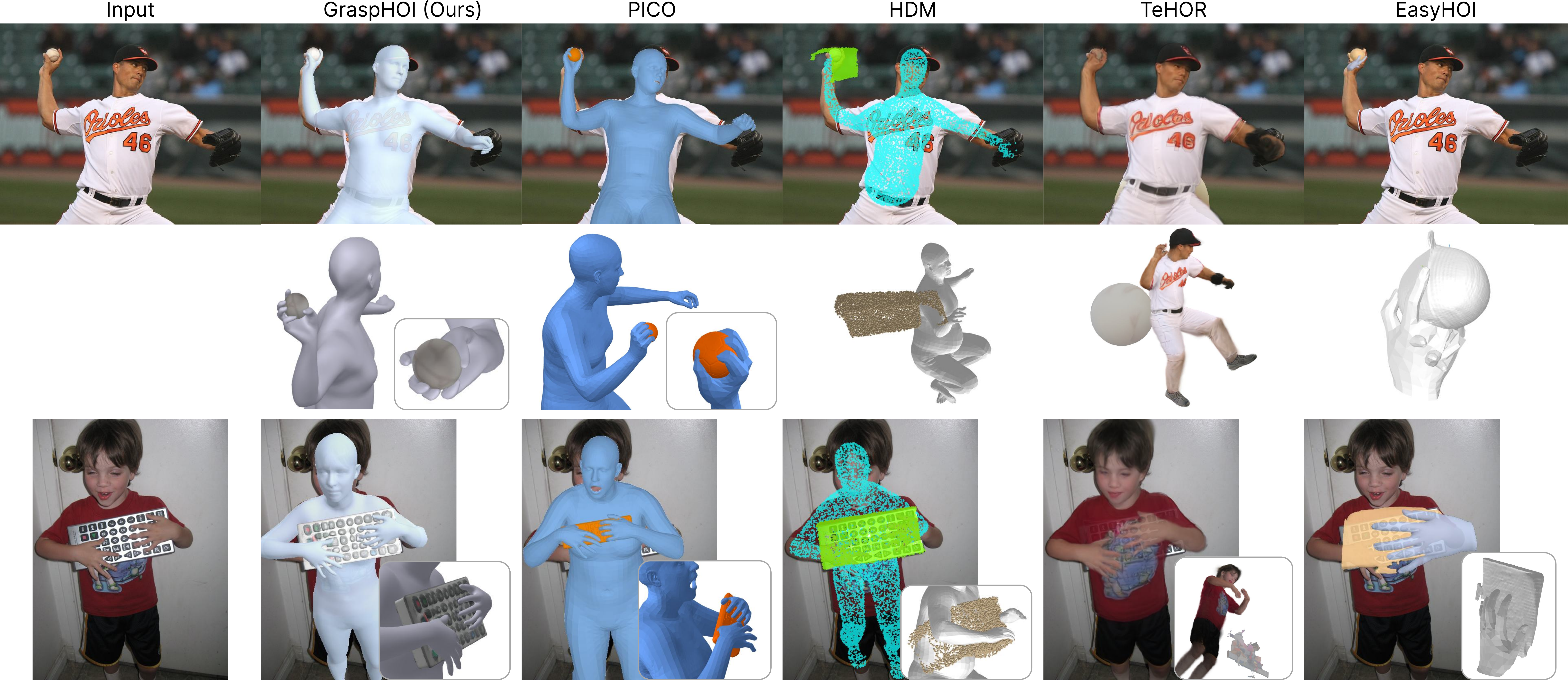}
  \caption{Qualitative comparison with PICO, HDM, TeHOR, and EasyHOI on
  challenging in-the-wild interactions. Method columns overlay the recovered
  human and object on the input, while insets expose their 3D arrangement and
  hand-object contact from alternate views. In these examples, GraspHOI better
  preserves the observed body pose, object geometry, scale, placement, and
  articulated grasp.}
  \label{fig:template_comp}
  \vspace{-1mm}
\end{figure*}

\subsection{HOI Baseline Comparison}
\label{sec:hoi_baselines}

We compare with HDM~\cite{xie2023template_free}
and TeHOR~\cite{Nam_2026_CVPR}, where HDM is category-bound but we use the ``general'' mode of its PC$^2$~\cite{melaskyriazi2023projection} reconstruction, and TeHOR is open vocabulary. We exclude CARI4D~\cite{xie2025cari4d}, whose video protocol needs temporal evidence and a large rigid object visible in the first frame. HDM, TeHOR, and GraspHOI are evaluated on matched ARCTIC and PGG subsets of 4{,}373 and 3{,}710 frames, respectively. For the signed-distance metrics, we mesh HDM's object point cloud with screened Poisson reconstruction.
After resolving HDM's global scale ambiguity, HDM and GraspHOI have similar
human sCD and MPJPE on PGG, while a large margin exists for ARCTIC. This is largely due to HDM's inability to handle occlusions, as PGG has no body occlusion due to the ProciGen setup, while ARCTIC constantly has the legs occluded due to a table. GraspHOI outperforms both methods in every hand metric, indicating better wrist position, better hand poses, and better contact overall. TeHOR's near-zero PD coincides with a large FL, as its
predicted hands usually remain far from the object rather than forming a grasp. In Table~\ref{tab:stagewise}, GraspHOI's foundation models without optimized placement, seating, and refinement reach 25.5\% Contact and 38.30\,mm PJO, versus 86.2\% and 13.30\,mm with them.

\subsection{Optimization-Based Methods}

PICO~\cite{Cseke2025PICO} and GraspHOI are evaluated on PICO-db using FL, PD, and Offset. We also evaluate
EasyHOI~\cite{Liu_2025_CVPR}, an optimization-based hand-object reconstruction
method, on ARCTIC and PGG. In these full-body images, EasyHOI struggles with
low-resolution hand and object evidence. Additionally, frames on which EasyHOI fails are removed from the evaluation set. Table~\ref{tab:optimization_baselines} uses 3.7K ARCTIC and 2.3K PGG frames. For PICO-db, 1,223 selected images that have hand contact are scored, excluding some large objects such as chairs and beds. Ours outperforms PICO by a large margin in both float and penetration. PICO receives the dataset's annotated contact maps as input, whereas GraspHOI infers
contact from the image, and yet our method provides more plausible contact. Even on the frames where it succeeds, and despite its internal hand cropping step, EasyHOI’s hands are less accurate, including after Procrustes alignment.

\begin{table}[!t]
\centering
\small
\setlength{\tabcolsep}{3.2pt}
\resizebox{\columnwidth}{!}{%
\begin{tabular}{llcccc}
\toprule
Dataset & Method & PA-MPJPE $\downarrow$ & Offset (FL / PD) $\downarrow$
& PJO $\downarrow$ & 2D MPJPE $\downarrow$ \\
\midrule
\multirow{2}{*}{PICO-db}
& PICO & - & 16.24 (11.90 / 4.34) & - & - \\
& GraspHOI & - & \textbf{4.56} (2.60 / 1.96) & - & - \\
\midrule
\multirow{2}{*}{ARCTIC}
& EasyHOI & 9.87 & 51.71 (48.15 / 3.56) & 72.44 & 288.79 \\
& GraspHOI & \textbf{5.69} & \textbf{2.18} (0.30 / 1.89) & \textbf{12.85} & \textbf{20.61} \\
\midrule
\multirow{2}{*}{PGG}
& EasyHOI & 13.29 & 43.12 (39.79 / 3.33) & 83.41 & 40.17 \\
& GraspHOI & \textbf{6.83} & \textbf{6.44} (4.73 / 1.71) & \textbf{17.17} & \textbf{7.33} \\
\bottomrule
\end{tabular}}
\caption{Optimization-based comparison. PA-MPJPE, FL, PD, Offset, and PJO
are in mm, and 2D MPJPE is in pixels. PICO-db lacks posed 3D hand-joint ground truth.}
\label{tab:optimization_baselines}
\vspace{-2.5mm}
\end{table}

\subsection{Qualitative Results}

Figure~\ref{fig:template_comp} compares representative reconstructions. PICO~\cite{Cseke2025PICO} shows large
body-pose and object-scale errors. Its retrieved contact pulls the contacting limb to a
retrieved mesh rather than the observed object. Although
HDM~\cite{xie2023template_free} uses a joint diffusion
prior, it struggles with body pose and object geometry, failing to recover
even the in-distribution ball and box-like shapes. TeHOR~\cite{Nam_2026_CVPR} recovers reasonable
body poses but produces large object-scale and placement errors. For the
occluded examples, its TRELLIS stage reconstructs strongly deformed geometry.
EasyHOI~\cite{Liu_2025_CVPR}
begins from plausible hand and object estimates, yet its optimization becomes
unstable when both occupy too few pixels to provide reliable refinement cues.
Figure~\ref{fig:teaser} further shows GraspHOI on diverse Internet images.

\subsection{Ablation Study}
\label{sec:ablation}

\begin{table*}[!t]
\centering
\small
\setlength{\tabcolsep}{4pt}
\begin{tabular*}{\textwidth}{@{\extracolsep{\fill}}cllll|ccc@{}}
\toprule
\# & Hands & Object pose & Object scale & Refined & Offset (FL / PD) $\downarrow$ & PJO $\downarrow$ & Contact (\%) $\uparrow$ \\
\midrule
1 & SAM 3D Body & Layout & Depth ($N_d$) & -- & 22.32 (18.12 / 4.20) & 37.80 & 22.5 \\
2 & WiLoR & Layout & Depth ($N_d$) & -- & 23.21 (19.57 / 3.64) & 38.30 & 25.5 \\
3 & WiLoR & Optimized & Depth ($N_d$) & -- & 20.54 (17.36 / 3.19) & 34.60 & 30.8 \\
4 & WiLoR & Optimized & Contact ($N^*$) & -- & 7.66 (3.95 / 3.72) & 18.10 & 35.4 \\
5 & WiLoR & Optimized & Contact ($N^*$) & Finger only & 5.39 (3.33 / 2.06) & 17.42 & 61.9 \\
6 & WiLoR & Optimized & Contact ($N^*$) & Wrist + finger & 3.75 (1.92 / 1.83) & 15.26 & 75.2 \\
7 & WiLoR & Optimized & Contact ($N^*$) & Full arm + finger & \textbf{2.14} (0.34 / 1.80) & \textbf{13.30} & \textbf{86.2} \\
\bottomrule
\end{tabular*}
\caption{Stage-wise ablation on the ARCTIC hands of Tables~\ref{tab:hoi_baselines} and~\ref{tab:joint_ablation}. Each row adds one stage. All metrics except Contact are in mm.}
\label{tab:stagewise}
\end{table*}

\begin{figure}[!t]
\centering
\includegraphics[width=\columnwidth]{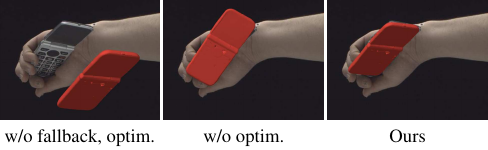}
\caption{Object-placement ablation on an ARCTIC frame. The placed object (red)
is projected onto the input image before joint refinement.}
\label{fig:object_ablation}
\vspace{-1mm}
\end{figure}

\textbf{Object placement.}
We ablate the object branch on the ARCTIC set, showing the importance of our fallback mechanism and object optimization. Contour (px) is the mean symmetric distance between the projected and ground-truth object contours at native resolution.
Table~\ref{tab:object_ablation} shows that \emph{w/o fallback, optim}
places the object by the SAM 3D Objects layout as predicted but is often misplaced. \emph{w/o optim} adds the FoundationPose registration fallback, but is not yet properly aligned with the input image. Our full method improves object placement overall. All variants are
scaled to metric depth as in Sec.~\ref{sec:method_obj} and share every other
stage. Fig.~\ref{fig:object_ablation} shows an example of each step.

\begin{table}[!t]
\centering
\small
\setlength{\tabcolsep}{3.2pt}
\begin{tabular*}{\columnwidth}{@{\extracolsep{\fill}}lccc@{}}
\toprule
Object placement & Object sCD $\downarrow$ & IoU $\uparrow$ & Contour (px) $\downarrow$ \\
\midrule
w/o fallback, optim. & 42.37 & 0.584 & 33.26 \\
w/o optim. & 39.79 & 0.681 & 23.66 \\
Ours & \textbf{36.71} & \textbf{0.800} & \textbf{12.20} \\
\bottomrule
\end{tabular*}
\caption{Object-placement ablation on 5.2K ARCTIC frames, before joint refinement. Object sCD is as in Table~\ref{tab:hoi_baselines}; IoU and Contour use the ground-truth silhouette.}
\label{tab:object_ablation}
\vspace{-1mm}
\end{table}

\begin{table}[!t]
\centering
\small
\setlength{\tabcolsep}{3.2pt}
\begin{tabular*}{\columnwidth}{@{\extracolsep{\fill}}lccccc@{}}
\toprule
Joint refinement & FL & PD & Offset $\downarrow$ & PJO $\downarrow$ & Contact (\%) $\uparrow$ \\
\midrule
w/o $\mathcal L_{\mathrm{pen}}$ & 0.26 & 5.69 & 5.95 & 15.19 & 32.7 \\
w/o $\mathcal L_{\mathrm{cont}}$ & 23.20 & 0.61 & 23.80 & 42.05 & 50.9 \\
Ours & 0.34 & 1.80 & \textbf{2.14} & \textbf{13.30} & \textbf{86.2} \\
\bottomrule
\end{tabular*}
\caption{Joint-refinement ablation on ARCTIC. All metrics except Contact are in mm.}
\label{tab:joint_ablation}
\vspace{-1mm}
\end{table}

\begin{figure}[!t]
\centering
\includegraphics[width=\columnwidth]{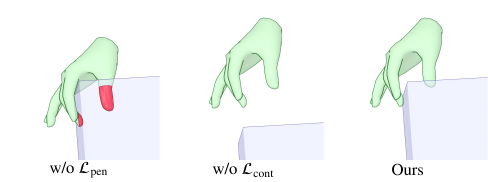}
\caption{Effect of the penetration and contact losses during joint refinement. Red regions indicate hand-object intersection, where intersection volume is 8.70~cm$^3$ without $\mathcal L_{\mathrm{pen}}$ and zero for ours, and without $\mathcal L_{\mathrm{cont}}$ the hand floats 5.1~cm above
the box.}
\label{fig:ablation2}
\vspace{-2.5mm}
\end{figure}

\textbf{Contact modulation.}
Reconstructed objects may be locally too thick near the
fingers, causing penetration even after coarse placement. Unsigned contact
attraction neither distinguishes interior from exterior pairs nor explicitly
pushes fingers outward, often settling into local minima. Our signed
penetration loss reduces these intersections (Fig.~\ref{fig:ablation2}), though
occasionally at the cost of agreement with the input finger pose.
Table~\ref{tab:joint_ablation} ablates the two contact terms of joint
refinement. Without $\mathcal L_{\mathrm{cont}}$, the hands do not stay in contact with the object surface and often drift away, explaining the large FL value. Without $\mathcal L_{\mathrm{pen}}$, fingers sink into the object, with the highest PD. With both terms, GraspHOI attains the
lowest Offset, PJO, and highest Contact (\%), indicating that our method keeps most hands on the surface.

\textbf{Stage-wise ablation.}
Table~\ref{tab:stagewise} adds the stages of GraspHOI one at a time, and its rows 4--7 vary which joints the refinement may rotate. Row 4
is the seated result before joint refinement. Following rows run the same
refinement for fingers only, then wrist + finger, and finally the full arm. Each added joint group lowers Offset and raises Contact (\%), and refining the full arm keeps the most hands on the surface. Refining the fingers mainly lowers PD, whereas freeing the wrist and then the shoulder and
elbow mainly lowers FL. Contact therefore is heavily influenced by where the arm places the hand, not only by finger articulation. This may raise joint error against the ground truth, a cost we accept for plausible contact, although the refined hands still have the lowest hand-pose errors in Tables~\ref{tab:hoi_baselines} and~\ref{tab:optimization_baselines}.

\section{Conclusion}

We introduced GraspHOI, a monocular framework that reconstructs a full-body 3D HOI and explicitly optimizes arm and finger articulation against the object while operating category agnostically. It targets a failure especially visible in animation, game cutscenes, and virtual reality: globally plausible interactions whose fingers float, penetrate the object, or remain in a generic pose rather than forming a grasp.

GraspHOI reconstructs the object, body, and hands separately and aligns their scale and placement in a shared metric frame. The final optimization can then focus on arm and finger articulation and surface contact instead of re-estimating the entire scene.

Across our evaluations, \mbox{GraspHOI} improves relative placement and contact plausibility. Upstream errors can still propagate through refinement, and the method reconstructs static interactions rather than motion. Our method is also sensitive to occlusion, which reduces the visible evidence for contact. Failure cases, limitations, and future work are detailed in the supplementary material. Its plausible, category-agnostic outputs can serve as assets or keyframes for HOI video generation.

\nocite{carion2025sam,rombach2021highresolution}
{
    \small
    \bibliographystyle{ieeenat_fullname}
    \bibliography{main}
}

%%%%%%%%% SUPPLEMENTARY MATERIAL (same numbering as the separate supplement: S sections, figures, tables, equations)
\clearpage
\setcounter{section}{0}
\setcounter{figure}{0}
\setcounter{table}{0}
\setcounter{equation}{0}
\renewcommand{\thesection}{S\arabic{section}}
\renewcommand{\thesubsection}{\thesection.\arabic{subsection}}
\renewcommand{\thefigure}{S\arabic{figure}}
\renewcommand{\thetable}{S\arabic{table}}
\renewcommand{\theequation}{S\arabic{equation}}
\renewcommand{\theHsection}{S\arabic{section}}
\renewcommand{\theHfigure}{S\arabic{figure}}
\renewcommand{\theHtable}{S\arabic{table}}
\renewcommand{\theHequation}{S\arabic{equation}}

\twocolumn[
\begin{center}
{\LARGE\bfseries GraspHOI: Full-Body 3D Human-Object Reconstruction with\\
Articulated Hands from a Single In-the-Wild Image\\[3pt]
Supplementary Material\par}
\end{center}

]

\hyphenation{Foundation-Pose ProciGen ARCTIC Procrustes EasyHOI GraspHOI CameraHMR InterTrack InterCap}
This supplementary material documents implementation and evaluation details
omitted from the main paper for space. We specify the model conversions,
amodal conditioning, metric placement, contact association and seating,
optimization schedules, dataset preparation, baseline preprocessing, and
evaluation metrics. We also present extended qualitative results,
representative failures and limitations of the current pipeline, future work,
and licensing details for the
images shown.

\section{Additional Design Details}
\label{sec:supp_design}

\subsection{Contact modeling}
Our method assumes that the target interaction contains \textbf{at least one grasping
hand}. HOI-DETR~\cite{darkhalil2026improving} predicts
hand-object links, which we associate with left and right
SMPL-H~\cite{Romero2017MANO} hands and with the target object mask. Amodal
overlap then rejects geometrically inconsistent links. For retained contacts,
palmar ray intersections and the visible-object
occlusion order determine fixed one-to-one hand-object pairs. This separates
semantic contact detection from geometric contact localization.

\subsection{Input and hyperparameters}

Our method requires a visible object mask and an object-name prompt in addition
to the RGB image. The prompt is used only by the open-vocabulary inpainting
stage and never selects object geometry or contact.
Human boxes, hand boxes, and contact decisions are inferred automatically. No
human mask or contact map is supplied. Our method does not support multi-human settings.

The same hyperparameters are used across all datasets.
Section~\ref{sec:Sup_details} reports the optimization schedules and fallback
behavior used in our experiments.

\subsection{MHR to SMPL-H conversion}
SAM 3D Body~\cite{Yang_2026_CVPR} is run with frozen weights in its body-only
inference setting. It returns a camera-space MHR~\cite{ferguson2025mhr} mesh.
We follow the official MHR conversion procedure, retaining its MHR-to-SMPL
surface mapping and its staged fitting in the SMPL-X parameterization but
replacing the destination with a neutral SMPL-H model. This substitution is possible because SMPL and SMPL-H
share the 6,890-vertex body topology~\cite{loper2015smpl}, while SMPL-H exposes the body-and-hand
pose interface expected by the fitter. The resulting parameters provide global
orientation, the 21-joint body pose, shape, and translation. We set the
fitted finger articulation aside and insert neutral hand poses before the
independent WiLoR initialization described below. Only a hand without an
associated HOI-DETR box, or one that the first safeguard of
Sec.~\ref{sec:supp_hands} rejects, keeps the fitted articulation, which in this
body-only setting is an almost constant, relaxed hand. Thus, SAM 3D Body
supplies the full-body and camera-space initialization, while the finger pose
of every other hand comes from the hand-focused estimator.

\subsection{MANO to SMPL-H pose transfer}
WiLoR~\cite{potamias2024wilor} predicts MANO~\cite{Romero2017MANO} global
orientation and 15 articulated finger rotations.
We insert these hand estimates into the fitted SMPL-H body. First, the elbow rotation is optimized
so that the projected SMPL wrist agrees with the WiLoR wrist; this step poses
the body with the SMPL model~\cite{loper2015smpl}, whose body joints SMPL-H
shares. Let
$R_g,R_{sp},R_c,R_s,R_e$ be the SMPL-H global, spine, collar, shoulder, and elbow rotations, where $R_{sp}$ combines the three spine joints, and let
$R_h^{\mathrm{MANO}}$ be the global MANO hand orientation. The corresponding
local SMPL-H wrist rotation is
\begin{equation}
R_w=(R_g R_{sp} R_c R_s R_e)^{\top}R_h^{\mathrm{MANO}}.
\label{eq:supp_wrist_conversion}
\end{equation}
The MANO finger rotations are then inserted directly into the SMPL-H hand-pose
parameters. Finally, we refine $R_w$ by matching the projected SMPL-H finger
joints to the WiLoR keypoints. Thus, no vertices are copied between MANO and
SMPL-H and only pose parameters are transferred. Body shape remains fixed, and
the final mesh is generated by the standard SMPL-H forward pass.

\section{Details on Our Pipeline}
\label{sec:Sup_details}

This section specifies the input assumptions, initialization procedures,
fallbacks, and optimization schedules abbreviated in the main paper.

\subsection{Inputs and shared metric depth}

The pipeline receives an RGB image $I$, a binary mask $S_{\mathrm{vis}}$ of the
visible object, and an open-vocabulary object name used only to prompt amodal
inpainting. The object name never selects a mesh, category model, or pose
template. HOI-DETR~\cite{darkhalil2026improving} is applied to $I$ to obtain
hand boxes, first-object boxes (objects in direct contact with a hand), and
hand-object interaction links.

We run the MoGe-2-conditioned version of Pixel-Perfect Depth
(PPD)~\cite{xu2025pixel,wang2025moge2}. PPD produces a sharp relative depth map,
while MoGe-2 supplies metric depth and camera-aware semantic features. Following
the released PPD point-cloud procedure, the relative prediction is aligned to
the MoGe-2 metric depth with RANSAC. When this alignment saturates, \ie more than
$30\%$ of the valid pixels sit at the maximum depth, we use the MoGe-2 metric
depth directly. We then align this map to the reconstructed hands, where the
object is placed. Over the camera-facing vertices $v$ of the detected hands whose
projected pixels $x_v$ lie outside $S_{\mathrm{vis}}$, we compute
$\gamma=\operatorname{median}_{v}[v_z/D(x_v)]$ and apply $D\leftarrow\gamma D$
once. The two filters keep mostly hand surface: a back-facing vertex, or one
behind the visible object, would read the depth of another surface. With fewer than 20 such vertices, $\gamma$ is instead computed over every vertex of the initial body fit that projects into the image. The resulting map gives the metric scale of the placed object, the depth
observation of the fallback FoundationPose~\cite{wen2024foundationpose}
registration, and the depth of objects
without detected contact. Amodal masking uses only the depth order, which this
scaling does not change.
Camera intrinsics $K$ are estimated with HumanFoV, released with
CameraHMR~\cite{patel2024camerahmr}, and shared across all renderers.

\subsection{Occlusion-localized amodal conditioning}

Single-view completion is necessarily generative: it predicts a plausible
continuation rather than recovering unobserved ground truth. Our goal is
therefore to preserve the observed object interior and restrict generation to
a small region that can genuinely contain an occluder. This differs from amodal completion methods that synthesize the whole object from a partial view~\cite{ozguroglu2024pix2gestalt} or reconstruct it directly in 3D~\cite{wu2025amodal3r}.

Let $B_\delta(S_{\mathrm{vis}})$ be the visible-object bounding box expanded by
$\delta$, where $\delta$ is $25\%$ of the box diagonal and is clamped to
$[8,60]$ pixels. We estimate the far visible-object depth by
\begin{equation}
z_{\mathrm{far}}=P_{90}\{D(x):x\in S_{\mathrm{vis}}\}.
\end{equation}
Here $P_{90}$ denotes the 90th percentile of the valid depths in the set.
Pixels deeper than $z_{\mathrm{far}}$ are observed background and must not be
inpainted. The initial occluder region is therefore
\begin{equation}
S_{\mathrm{occ}}=
\left(B_\delta(S_{\mathrm{vis}})\setminus S_{\mathrm{vis}}\right)
\cap\{x:0<D(x)\leq z_{\mathrm{far}}\}.
\label{eq:supp_occluder}
\end{equation}
We dilate this mask only in a narrow band adjacent to the object and include a
two-pixel object-side rim at the object-occluder boundary. Stable Diffusion may
repaint this narrow rim to suppress color leakage from the occluder. Visible
pixels away from the rim remain fixed. This supplies enough support for the
generated continuation without allowing the object to grow into distant
background or invalid-depth regions.

\begin{figure*}[p]
  \centering
  \includegraphics[width=0.9\textwidth,height=0.86\textheight,keepaspectratio]{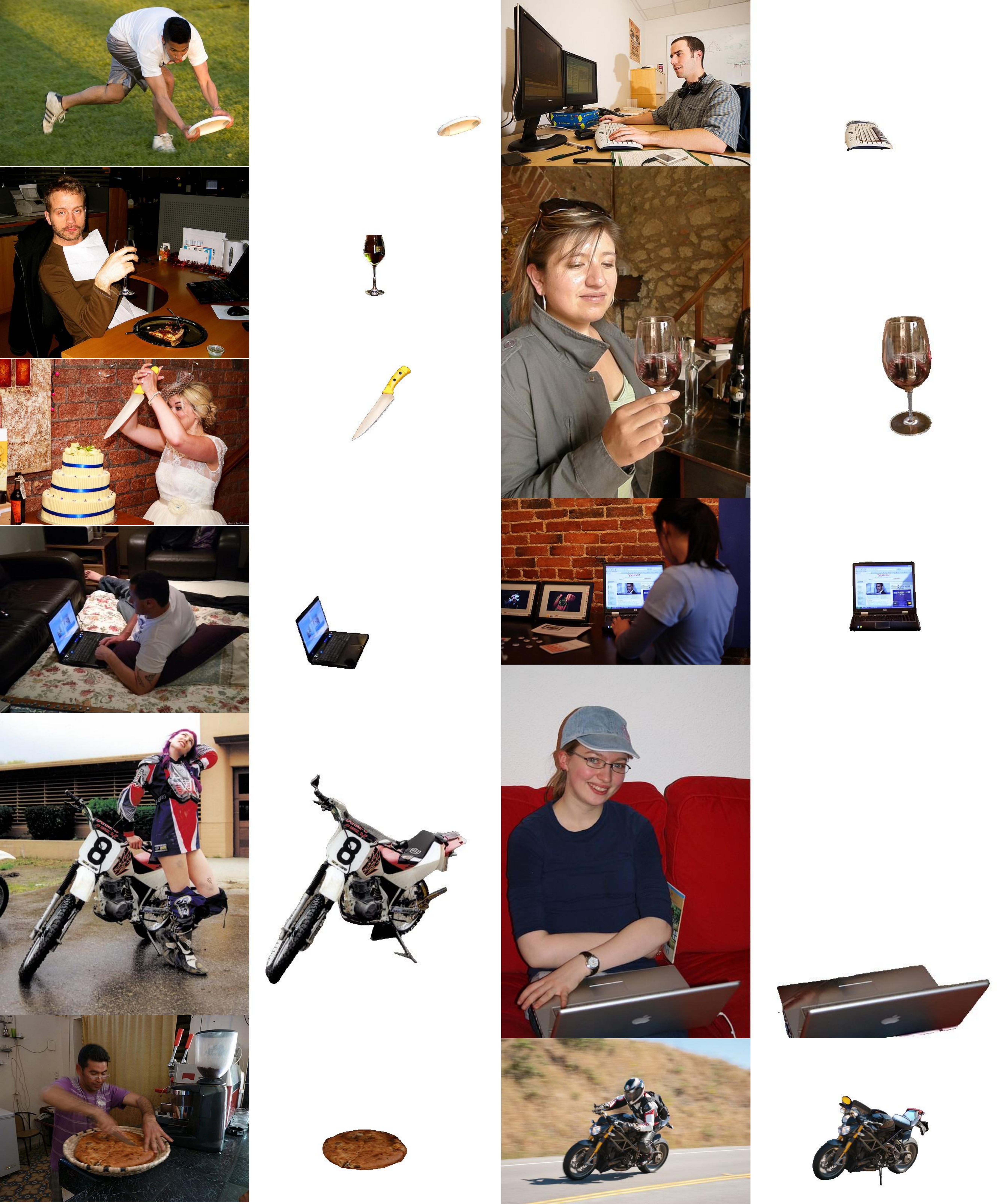}
  \par\smallskip\textbf{(a)}
  \caption{Examples of amodal conditioning on
  PICO-db. Across panels (a)--(c), each pair shows the input image and the extracted object cutout after occlusion-localized inpainting. Visible object support is retained while plausible object appearance is generated only in the inferred occluder region.}
  \label{fig:amodal}
\end{figure*}

We crop a square around $S_{\mathrm{vis}}\cup S_{\mathrm{occ}}$ and replace all
non-object pixels by a fixed gray texture. Stable Diffusion inpainting
\cite{rombach2021highresolution} receives this image, $S_{\mathrm{occ}}$, and
the prompt \emph{a [object name]}. We use the public Stable Diffusion 1.5 inpainting checkpoint. A foreground remover extracts the generated object from
the gray field, and its alpha is unioned with
$S_{\mathrm{vis}}$ so that visible object support cannot be discarded. This yields the amodal cutout $I_a$ and silhouette $S_a$ shown in Fig.~\ref{fig:amodal} (page~\pageref{fig:amodal}).

\paragraph{Validity of amodal completion.}
Every placement step fits the object to $S_a$, so $S_a$ must follow the true
object outline, including its occluded parts.
Table~\ref{tab:supp_amodal_validity} compares $S_a$ with the silhouette of the
SAM 3D Objects layout as predicted, the first row of
Table~\ref*{tab:object_ablation} of the main paper, against the ground-truth
object silhouette on the same ARCTIC frames. On average, $S_a$ covers 97.1\% of
the ground-truth silhouette, and 91.6\% of its pixels lie on the object. Its IoU
is higher than that of the layout on 99.7\% of the frames, and it falls below
0.5 on 1.5\% of the frames, against 27.9\% for the layout. $S_a$ is therefore a
reliable placement target, whereas the predicted layout alone is often
misaligned.

\begin{table}[t]
\centering
\small
\setlength{\tabcolsep}{3.2pt}
\begin{tabular*}{\columnwidth}{@{\extracolsep{\fill}}lcccc@{}}
\toprule
Silhouette & IoU $\uparrow$ & Precision $\uparrow$ & Recall $\uparrow$ & F1 $\uparrow$ \\
\midrule
Layout only & 0.584 & 0.657 & 0.791 & 0.712 \\
Ours ($S_a$) & \textbf{0.893} & \textbf{0.916} & \textbf{0.971} & \textbf{0.939} \\
\bottomrule
\end{tabular*}
\caption{Validity of amodal completion on ARCTIC, measured against the
ground-truth object silhouette.}
\label{tab:supp_amodal_validity}
\end{table}

\subsection{Human-hand assembly}
\label{sec:supp_hands}

HOI-DETR hand boxes are assigned to left and right by overlap with projected
fitted SMPL-H hand boxes. WiLoR is then run on each associated crop. The first of the two alignment stages uses Adam for 60 iterations at learning rate $10^{-2}$ to rotate the SMPL elbow until the projected wrist matches the WiLoR wrist.
After converting the MANO global orientation into the local wrist frame
(Eq.~\eqref{eq:supp_wrist_conversion}) and
transferring MANO finger rotations to SMPL-H, we optimize only the wrist
rotation for 100 iterations at learning rate $3\times10^{-2}$. The latter loss
is the squared reprojection error of the 15 SMPL-H finger joints against WiLoR
2D keypoints. The SMPL shape coefficients are fixed throughout.

Three safeguards keep the transferred wrist plausible. First, WiLoR can misread
a largely hidden hand, \eg one folded back over the forearm behind a cup. For
each hand, we measure the angle between the directions in which the
transferred WiLoR hand and SAM 3D Body's own hand point, from the wrist to the
middle-finger knuckle. Roll about this direction, \ie pronation and supination,
does not change it. If the angle exceeds $90^\circ$, we keep SAM 3D Body's wrist
and finger rotations for that hand. Without trusted keypoints, the hand is then
excluded from the keypoint and joint refinements. Second,
the keypoint refinement keeps each wrist rotation within $30^\circ$ of its
transferred value by projecting it back onto this cone after every step.
Third, SMPL-H skins the whole forearm twist at the wrist joint, which pinches
the wrist into a bow-tie shape at large twists. Where the twist of a wrist
about the forearm axis exceeds $60^\circ$, we move half of it to the elbow. This
leaves the hand's global rotation and all joint positions unchanged and turns
only the forearm skin.

\subsection{Object reconstruction and placement}
\label{sec:supp_obj}

\paragraph{Reconstruction.}
SAM 3D Objects~\cite{sam3dteam2025sam3d3dfyimages} receives the image with
$I_a$ pasted into $S_a\setminus S_{\mathrm{vis}}$ and the mask $S_a$, and
returns a textured mesh with its layout. The mesh is a closed two-layer crust:
an inner lining lies just under the outer surface, either as a separate
inverted shell with negative volume or joined to the outer surface through
small holes. In both cases, a winding-number inside-outside test reads the
object's interior as outside. We drop inverted shells and small fragments, decimate the mesh to about 40K faces, and delete the faces that no ray from outside reaches, which removes a joined lining. Each remaining boundary loop is
then capped by a triangle fan, and the faces are oriented outward, so that the
inside-outside tests of the contact seating and the joint refinement
(Secs.~\ref*{sec:method_contact} and~\ref*{sec:method_jointopt} of the main paper)
are well defined. The result is the mesh $\mathcal M_o^0$ of
Sec.~\ref*{sec:method_pre} of the main paper.
Vertex colors are carried over and are used by FoundationPose in the fallback.

\paragraph{Layout and translation re-fit.}
SAM 3D Objects predicts the layout in the camera of its MoGe depth estimator,
up to scale. We re-express it in $K$ by rotating the posed mesh about the
camera origin, so that its bounding-box center lies on the ray of the same
pixel under $K$, and by setting its distance so that its image footprint keeps
its area at that pixel. Shape and orientation relative to the line of sight are
unchanged. Each step of the translation re-fit places the bounding-box center
$c$ of the mesh at $\rho=\sqrt{|\widehat S|/|S_a|}$ times its depth, on the ray
through the pixel $\mu(S_a)-(\mu(\widehat S)-\Pi_K(c))/\rho$, where $\widehat S$
is the silhouette rendered under $K$, $\mu(\cdot)$ and $|\cdot|$ are the
silhouette centroid and area, and $\Pi_K(c)$ is the projection of $c$. Under
weak perspective, scaling the depth by $\rho$ shrinks the silhouette by
$1/\rho$ about $\Pi_K(c)$, so this pixel puts the new centroid exactly on
$\mu(S_a)$, and the area matches as well.
Under full perspective, the error left by a step is the correction times a
factor of the order of the object's depth range over its distance, so repeating
the step converges geometrically to the pose whose silhouette has the centroid
and area of $S_a$. A step is accepted only if it raises the IoU with
$S_a$. The render covers a canvas padded by half the
longer image side on each side, so that the part of the object outside the
frame is counted. When $S_a$ touches the image border, \ie the object is cut by the
frame, the clipped silhouette no longer follows the depth law, so the loop also
runs on the in-frame part of the silhouette, and the higher IoU of the two runs
is kept.

\paragraph{Gate and selection.}
FoundationPose registration is the slowest placement step, so a re-fitted layout
with an IoU of at least $0.7$, a common threshold for a confident overlap, is
accepted without it, unless it covers no valid depth pixels, so that its metric
scale cannot be measured. Otherwise, both the layout and the FoundationPose
registration are scaled and refined, and the placement with the higher final
IoU is kept. Because the better placement is kept, the threshold only trades
runtime for accuracy: a higher value can only raise the final IoU with $S_a$. Within either
path, the silhouette refinement is kept only if it raises the IoU of its
starting placement.
Over all processed images, the FoundationPose fallback runs on
34.4\% of the PICO-db, 15.3\% of the ARCTIC, and 15.6\% of the PGG images.

\paragraph{Metric scale.}
The scalar $N_d$ of Sec.~\ref*{sec:method_obj} of the main paper is the
median of $D(x)/z(x)$ over
the visible-object pixels $x$, where $z(x)$ is the rendered depth of $V_o$ and
$S_{\mathrm{vis}}$ is eroded by 2\,px against depth edges. Small objects with
fewer than 20 such pixels use the uneroded visible mask and then $S_a$. For images with detected contact, the contact seating then sets the final depth from the hand-object pairs; $N_d$ keeps only a weak influence through the silhouette refinement before it, whose steps are taken in meters. Without detected contact,
no hand-object pair constrains the depth, so the object stays at the depth given
by the depth map: $N_d$, or, if its silhouette refinement was kept, a depth
re-estimated as the median of $D/z$ over $S_a$ in the refinement crop.

\paragraph{Fallback registration.}
For metric scale, we back-project every valid depth pixel $x=(x_1,x_2)$ inside
$S_{\mathrm{vis}}$ using the camera intrinsics:
\begin{equation}
\mathcal P=\left\{D(x)K^{-1}[x_1,x_2,1]^\top:
x\in S_{\mathrm{vis}}\right\}.
\label{eq:supp_scale_cloud}
\end{equation}
Mask-boundary errors can create distant points, so we apply robust depth
filtering followed by statistical spatial outlier removal before estimating
scale.

Let $c_{\mathcal P}$ and $c_m$ be the coordinate-wise medians of the filtered
cloud and generated mesh vertices, respectively. Their enclosing radii and the
initial scale are
\begin{equation}
\begin{aligned}
r_{\mathcal P}&=\max_{p\in\mathcal P}\|p-c_{\mathcal P}\|_2, &
r_m&=\max_{v\in V_o^0}\|v-c_m\|_2,\\
s_{\mathrm{sph}}&=r_{\mathcal P}/r_m, &
\widetilde V_o^0&=s_{\mathrm{sph}}V_o^0.
\end{aligned}
\label{eq:supp_sphere_scale}
\end{equation}
Unlike a projected bounding-box ratio, this statistic has no preferred image
axis, although incomplete visible geometry can still bias it. If too few valid
points remain, we instead match the amodal-mask extent at its median metric
depth to the generated mesh extent.

FoundationPose~\cite{wen2024foundationpose} requires RGB-D observations.
We run PPD on the completed object and robustly align this relative depth to
the scene depth over $S_{\mathrm{vis}}$, where both maps observe the same
surface. The aligned object depth is pasted
inside $S_a$, and the completed RGB is pasted into the corresponding image
region. FoundationPose registers the textured mesh to this
as-if-unoccluded RGB-D observation and initializes rotation and translation.
Thus, FoundationPose receives the sphere-scaled mesh rather than estimating its
metric scale. The registered pose then passes through the same translation
re-fit, metric scale, and silhouette refinement as the layout. If FoundationPose
fails, the silhouette refinement instead starts from a mask-centered
initialization at the fingertip depth of the detected hands, with its projected scale
matched to the object mask, and its result is used without the re-fit, the
metric scale, or the IoU check. The refinement also falls back to this
initialization whenever its starting pose, on either path, is implausible: a
depth below 5\,cm, or a box-size correction outside $[1/4, 4]$.

\paragraph{Silhouette refinement.}
For stable boundary gradients on small objects, the silhouette refinement uses a
mask-centered square crop resized to $384^2$ pixels, with focal length and
principal point transformed to the crop camera. We optimize the object transform
with Adam at learning rate $3\times10^{-3}$ for at most 300 iterations, and stop
once the objective has not improved for 75 iterations, but not before iteration
150. The contour weight $w_c$ increases linearly from $0.3$ to $1.0$ over the 300
iterations, so an early-stopped fit ends at a lower value. The remaining weights
are exactly those of the object objective in
Eq.~(\ref*{eq:object_total_loss}) of the main paper. At the starting
pose, a geometric-mean correction of the rendered-to-observed box width and
height initializes the differentiable scale, and the translation is adjusted to
preserve the mesh center. We restore the iterate with the lowest objective
$\mathcal L_{\mathrm{obj}}$.

\subsection{Silhouette and contour losses}
\label{sec:supp_sil_cnt}

\begin{figure}[t]
\centering
\includegraphics[width=\columnwidth]{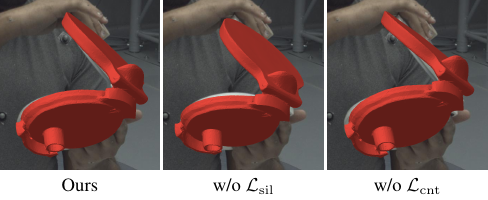}
\caption{An example showing the effect of each loss on an ARCTIC frame. The placed object (red)
is projected onto the input image before joint refinement.}
\label{fig:supp_sil_cnt}
\end{figure}

Table~\ref{tab:supp_sil_cnt} ablates the silhouette and contour losses of the
object refinement on the frames of Table~\ref*{tab:object_ablation} of the main
paper. The final two rows of the table show that neither loss requires the other, and numerically they provide near-identical performance. However, both losses exist for edge cases where the reconstructed object does not faithfully represent the object in the image, which is consequently not explicitly shown through metrics. Fig.~\ref{fig:supp_sil_cnt} shows an example in which removing
$\mathcal L_{\mathrm{sil}}$ causes steps to find a local minimum where the contour fits but the object is over-rotated in the process of doing so, while removing $\mathcal L_{\mathrm{cnt}}$ moves the object to cover the silhouette but neglects the angle and thickness of the upper lid. This phenomenon is caused by the fact that reconstructed objects are not identical to GT objects. While the differences exist visually, numerically these items report near-identical IoU, Contour, and object sCD. Thus, we leave the combination of silhouette and contour losses as a design choice rather than a contribution.

\begin{table}[t]
\centering
\small
\setlength{\tabcolsep}{3.2pt}
\begin{tabular*}{\columnwidth}{@{\extracolsep{\fill}}lccc@{}}
\toprule
Object placement & Object sCD $\downarrow$ & IoU $\uparrow$ & Contour (px) $\downarrow$ \\
\midrule
Ours & 36.71 & 0.800 & 12.20 \\
w/o $\mathcal L_{\mathrm{sil}}$ & 36.68 & 0.792 & 12.75 \\
w/o $\mathcal L_{\mathrm{cnt}}$ & 36.81 & 0.802 & 12.00 \\
\bottomrule
\end{tabular*}
\caption{Silhouette and contour loss ablation on ARCTIC. All metrics are measured before joint refinement, and IoU and Contour use the ground-truth object silhouette.}
\label{tab:supp_sil_cnt}
\end{table}

\subsection{Contact association}

\begin{figure}[!tbp]
\centering
\includegraphics[width=\linewidth]{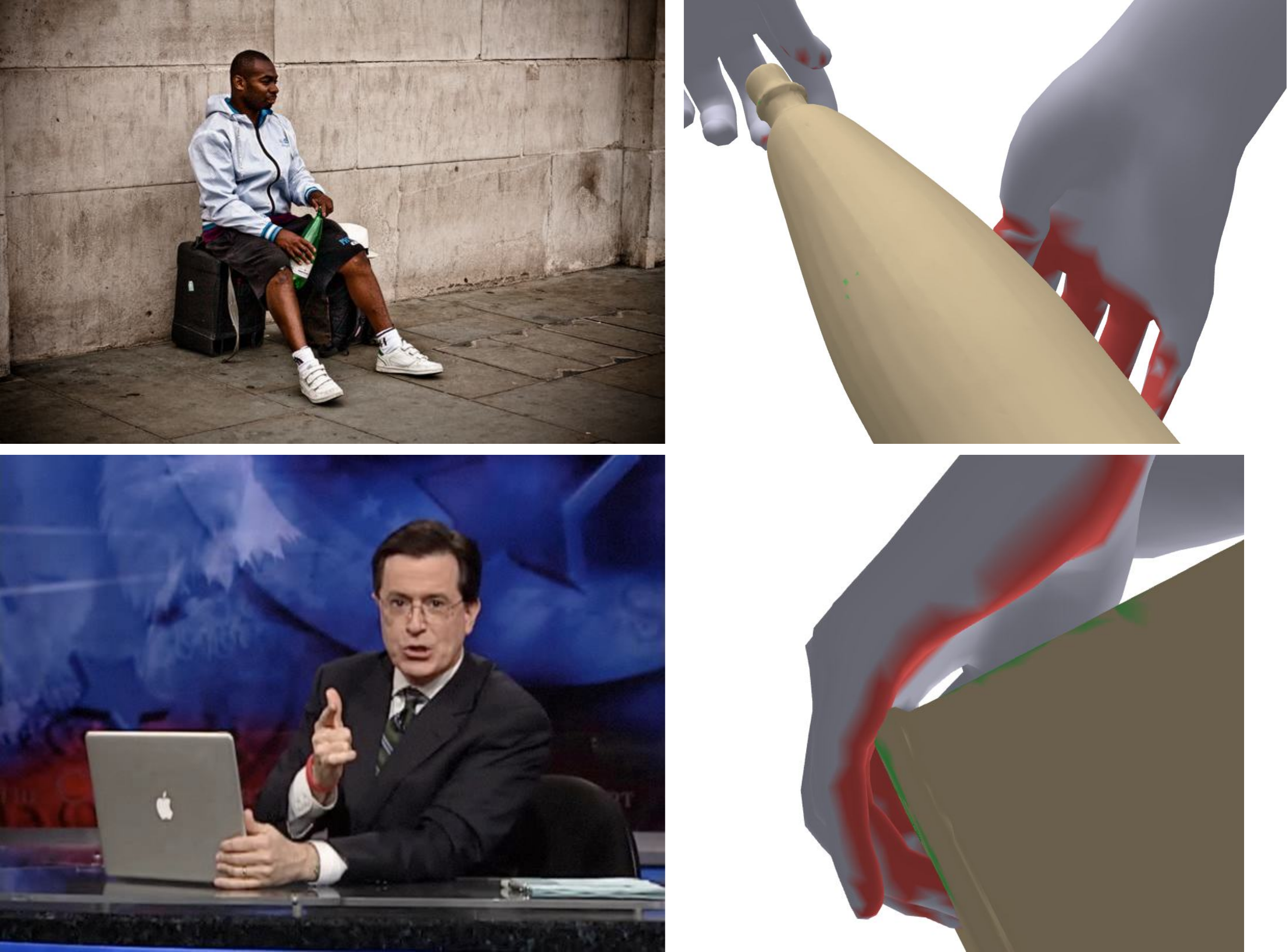}
\caption{Contact association examples. Red vertices are the hand points
where our ray detects contact as happening.}
\label{fig:contact_example}
\end{figure}

HOI-DETR predicts whether each detected hand is linked to a first object. We
match that object box to $S_a$ so that a hand manipulating a different object
does not become a false contact. Each HOI-DETR hand is associated with the
projected fitted SMPL-H left or right hand. As a geometric backstop, we demote a
predicted contact when the projected box of its SMPL-H hand overlaps fewer than
25 pixels of $S_a$.

For a retained contact, we use the full SMPL-H hand region rather than a small
set of fingertip or palm landmarks. Rays are cast through the intersection of
the hand bounding region and $S_a$. Every ray-triangle intersection is snapped
to its nearest hand vertex, after which the vertex normal is compared with the
mean palm normal. Candidates on the dorsal side are removed. For each remaining
palmar hand vertex, object vertices projecting to the same local ray are
collected within a 6-pixel radius. The visible mask determines the relevant
surface. A visible object pixel selects the far object surface, which faces a
hand behind the object. An occluded object pixel selects the near surface,
which faces a hand in front. Examples of this method are shown in Fig.~\ref{fig:contact_example}.

\paragraph{Occlusion-aware seating.}
The seated object is $NV_o$, where $N$ is searched on a one-dimensional grid.
The first pass minimizes the mean squared paired depth residual. If the seated
mesh still penetrates the hand by more than 2\,mm, a second pass adds an SDF
penetration penalty with weight $\lambda_{\mathrm{seat}}=20$. This penalty, $\mathcal L_{\mathrm{pen}}^{\mathrm{seat}}(N)$ in
Eq.~(\ref*{eq:depth_seating}) of the main paper, is the mean squared depth of
hand vertices inside the object: the mean of $[-\operatorname{SDF}_N(v)]_+^2$
over the vertices $v$ of the full SMPL-H hand regions of the contacting hands,
778 per hand, where $\operatorname{SDF}_N$ is the signed distance to the object
with vertices $NV_o$, positive outside, so vertices outside the object
contribute zero. Both passes use four-level coarse-to-fine grid
search. The first spans $[0.2N_0,5N_0]$ around the median hand/object depth ratio
$N_0$, and the penetration-aware pass spans $[0.5N_1,1.6N_1]$ around the first-pass
solution $N_1$. The final hand and object indices form fixed one-to-one
correspondences. When fewer than three reliable ray pairs are available,
seating uses representative fingertip vertices and nearest-surface distance instead.
Joint refinement likewise uses nearest-surface contact for any hand without
enough fixed pairs, ensuring that sparse ray intersections do not remove its
contact objective entirely.

\subsection{Contact-aware joint refinement}
\label{sec:supp_joint}

The object orientation is frozen after object placement. A learnable scalar $s_o\in[0.95,1.05]$ multiplies its camera-space vertices, \begin{equation} V_o'=s_oV_o. \end{equation} Here $V_o$ holds the seated camera-space object vertices $o_j$ of the main paper. Because perspective projection is
homogeneous, this changes object depth and metric size while leaving the
silhouette fixed. We define an SDF $\operatorname{SDF}_{\mathcal M_o}$, positive outside and negative inside, once for the unscaled object $\mathcal M_o$. For any camera-space point $p$, the scaled SDF of Eq.~(\ref*{eq:penetration_loss}) of the main paper is evaluated without rebuilding the grid: \begin{equation} \operatorname{SDF}_{s_o\mathcal M_o}(p)=s_o\operatorname{SDF}_{\mathcal M_o}\!\left(\frac{p}{s_o}\right). \label{eq:sdf_scaling} \end{equation}

The standard stage optimizes shoulder, elbow, wrist, and MANO finger rotations, together with $s_o$, for 500 Adam iterations at learning rate $3\times10^{-3}$. Following Eq.~(\ref*{eq:joint_total_loss}) of the main paper, its default objective is \begin{equation} \begin{aligned} \mathcal L_{\mathrm{joint}}={}&0.01\mathcal L_{\mathrm{kp}}+ 1.5\times10^5(\mathcal L_{\mathrm{pen}}+30\mathcal L_{\mathrm{tr}})\\ &+100\mathcal L_{\mathrm{reg}}+ 3\times10^4\mathcal L_{\mathrm{cont}}. \end{aligned} \end{equation}
Contact is enforced without a margin, while the
penetration compliance margin is 2\,mm. The 2D errors of $\mathcal L_{\mathrm{kp}}$ are measured in pixels of the input image.

\paragraph{Pose regularization.}
Eq.~(\ref*{eq:regularization_loss}) of the main paper holds each refined joint
$k$ near its initial rotation using two terms, where $d_k$ is the geodesic angle
between the joint's current and initial rotations; $d_k$ and $c_k$ enter every formula in radians, although we quote $c_k$ in degrees. The first term,
$\omega_k d_k^2$, pulls the joint back toward its initial rotation, and
$\omega_k$ sets the strength of this pull. The second term,
$\kappa_k[d_k-c_k]_+^{p_k}$, is zero until the joint has turned by the limit
angle $c_k$ and penalizes any rotation beyond it, with strength $\kappa_k$ and
power $p_k$.

The pull has a total weight of $0.5$ for the three joints of an arm and $0.3$
for the 15 finger joints of a hand, averaged over those joints, so
$\omega_k=0.5/3\approx0.167$ for arm joints and $\omega_k=0.3/15=0.02$ for finger
joints. An arm joint is therefore pulled about eight times harder than a finger
joint. The fingers need to be flexible enough to touch the object, whereas the
arms should stay in place if possible, as their movement breaks realism greatly.
The same rotation also moves the fingertips much farther at an arm joint.

For the second term, finger joints use $c_k=3^\circ$, $p_k=4$, and
$\kappa_k=20\,\omega_k/c_k^4$. A rotation of $3^\circ$ moves a fingertip by only a
few millimeters, about the compliance margin, so a finger joint can settle onto
the surface but cannot re-pose the hand. Past this limit the term quickly
outweighs the pull.
Arm joints instead use the rotation caps of item (iii) below.

When both hands contact the object, a coarse two-arm stage precedes the standard stage. It optimizes both shoulder-elbow-wrist chains for 500 iterations at learning rate $10^{-2}$, with keypoint, contact, penetration, and regularization weights $(1,10^4,5\times10^5,100)$. The coarse stage averages pair distances, whereas the standard stage uses the squared pair distances of Eq.~(\ref*{eq:contact_loss}) of the main paper. Finger rotations
remain fixed during the coarse stage. The standard stage then updates both arms, both MANO hand poses, and the bounded object-depth scalar using the same fixed contact pairs.

\paragraph{Safeguards.}
Both stages use three safeguards: a tunneling term (i) inside $\mathcal L_{\mathrm{pen}}$, plus a 2D trust region (ii) and arm rotation caps (iii) that keep the stronger penetration
weights from trading image evidence for contact.
(i) Thin objects such as cup walls, frisbees, and book pages leave the SDF
almost no interior to push a finger out of, and past the mid-plane of a slab
its gradient pushes a fingertip out the far side. At every iteration, we
therefore ray-test each bone of each finger chain, from the wrist to the
fingertip, against the object. For the first bone that crosses the surface, the
joint just past the crossing is pulled back to the wrist side of the crossed
triangle's plane with a 2\,mm margin, by a squared hinge on its signed distance
to that plane:
\begin{equation}
\mathcal L_{\mathrm{tun}}=\sum_{f\in\mathcal F_\times}
\left[m-n_f^{\top}(x_f-c_f)\right]_+^2,
\label{eq:supp_tunnel}
\end{equation}
where $\mathcal F_\times$ is the set of fingers with a bone that crosses the
surface, $c_f$ is the first crossing from the wrist, $n_f$ is the crossed
triangle's normal oriented toward the wrist, $x_f$ is the joint just past the
crossing, and $m=2$\,mm. Only this joint is pulled, and the next bone is tested again at
the next iteration. Pulling every joint past the crossing behind one plane
instead unwrapped fingers that legitimately curl around curved objects.
(ii) A 2D trust region $\mathcal L_{\mathrm{tr}}$ penalizes the squared drift of each hand joint beyond 1\,cm, and of each shoulder and elbow beyond 2\,cm, from its initial projection, with the pixel drift converted to meters at the joint's depth. Term (i) enters $\mathcal L_{\mathrm{pen}}$ with weight 10 (Eq.~(\ref*{eq:penetration_loss}) of the main paper); (ii) enters Eq.~(\ref*{eq:joint_total_loss}) of the main paper as $30\mathcal L_{\mathrm{tr}}$, scaled by $\lambda_{\mathrm{pen}}$ together with $\mathcal L_{\mathrm{pen}}$.
The \emph{w/o $\mathcal L_{\mathrm{pen}}$} ablation of
Table~\ref*{tab:joint_ablation} of the main paper removes $\mathcal L_{\mathrm{pen}}$, \ie the SDF term and (i), and keeps (ii).
(iii) The shoulder, elbow, and wrist rotations are softly capped at $15^\circ$,
$20^\circ$, and $30^\circ$ from their initial values by a squared hinge on the
geodesic angle, weighted by $\kappa_k=10^3$ within $\mathcal L_{\mathrm{reg}}$. This
keeps the elbow from swinging and the wrist from twisting into the bow-tie
shape described in Sec.~\ref{sec:supp_hands}.
Every iteration's state is scored with these weights, and
the best-scoring state is restored at the end.
To save runtime, either stage stops once this score has not improved for 100 iterations, but only after iteration 250, and not while $\mathcal L_{\mathrm{pen}}+30\mathcal L_{\mathrm{tr}}$ of the best state exceeds $10^{-6}$\,m$^2$, the square of 1\,mm.

\begin{figure*}[!t]
\centering
\includegraphics[width=\textwidth]{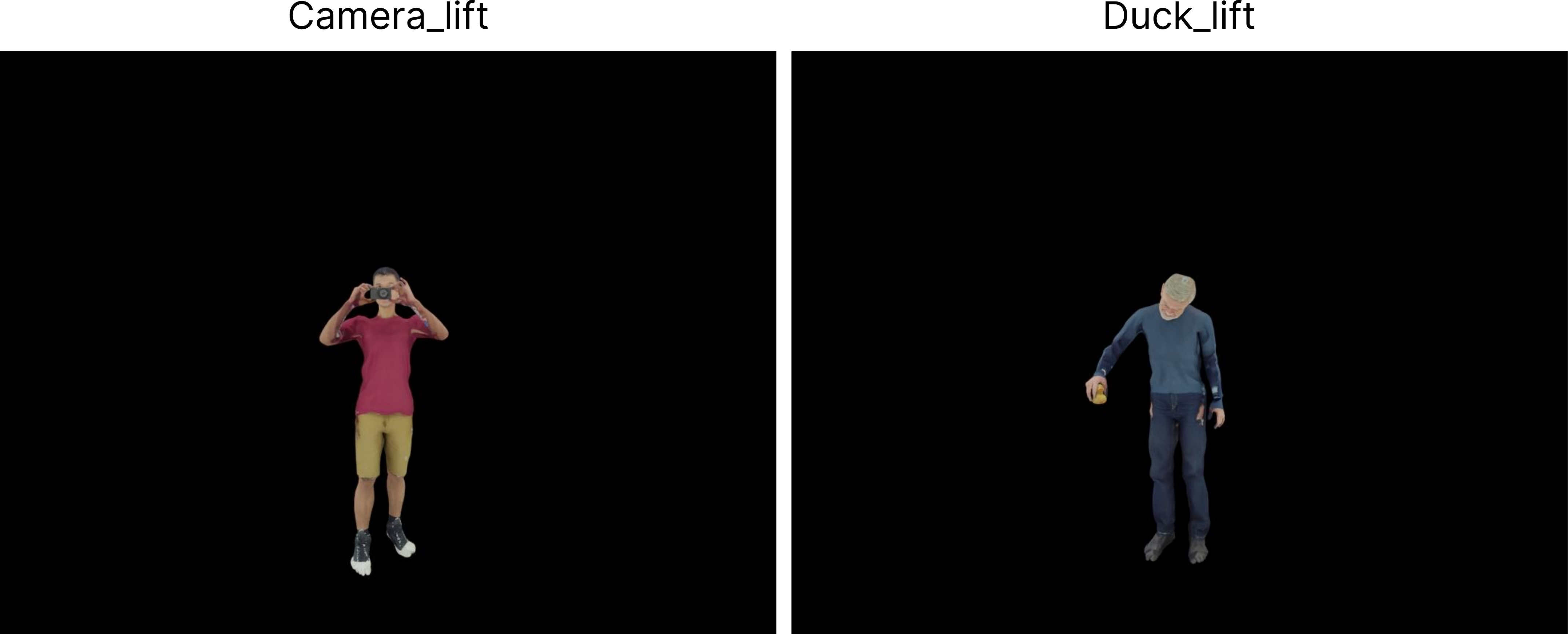}
\caption{ProciGen-GRAB examples. The small people, hands, and objects in the
frame are an intentional part of the dataset design.}
\label{fig:pgg_examples}
\end{figure*}

\subsection{Stage-wise ablation}
\label{sec:supp_stagewise}
Table~\ref*{tab:stagewise} of the main paper adds the stages of GraspHOI one at a time on the hands of
Table~\ref*{tab:joint_ablation} of the main paper. In Table~\ref*{tab:stagewise} of the main paper, \emph{Hands} are SAM~3D~Body's native hands from its body-only setting or the WiLoR transplant, and \emph{Object pose} is the SAM~3D~Objects layout or our optimized placement (translation re-fit, FoundationPose fallback, and silhouette refinement). Since the layout is not metric, \emph{Object scale} is the depth-based $N_d$ or the contact-based $N^*$ of occlusion-aware seating (Secs.~\ref*{sec:method_obj} and~\ref*{sec:method_contact} of the main paper), and \emph{Refined} lists the joint groups that refinement may rotate. All rows use the amodally conditioned reconstruction (Sec.~\ref*{sec:method_pre} of the main paper). As we claimed in the main paper, raw placement of foundation models is insufficient to reconstruct interactions. Rows 1 and 2 compare SAM 3D Body native hands with WiLoR hands. Our optimization steps increase contact performance.

Because our object is in a predicted shape and pose, we could achieve 0\,mm MPJPE and still have a high Offset and a low Contact (\%). Thus, as we target plausible rather than joint-accurate hands, we do not report MPJPE metrics in this ablation.

We argue that small deviations in hand contact quickly break realism, and to reinforce contact, it is necessary to move the arms toward the object as the predicted mesh is not identical to ground truth, and monocular settings inherently have depth ambiguities.

Contact (\%) supports this choice. To compare with the ground truth, we keep the 3.5K hands of Table~\ref*{tab:stagewise} of the main paper that are in contact in the ground truth. \textbf{Under the same 3.5K hands for all percentages}, the ground truth reaches 97.4\% Contact and ours reaches 86.3\% (86.2\% on all hands of Table~\ref*{tab:stagewise} of the main paper). Freeing the shoulder and elbow halved the gap, from 22 to 11 percentage points. The ground truth stays below 100\% because of differences in definitions: 2.6\% of its annotated hands do not meet our 3\,mm Offset threshold for Contact (\%), all of them penetrating deeper than 3\,mm on average.

\section{Dataset and Baseline Selection}

\subsection{ProciGen-GRAB}
GRAB~\cite{Taheri2020GRAB} provides metric
SMPL-X~\cite{pavlakos2019smplx} bodies, articulated hands,
object trajectories, and contact annotations, but not captured RGB. We construct
ProciGen-GRAB (PGG) by replacing the original source
interactions in the ProciGen~\cite{xie2023template_free} renderer with GRAB interactions
while retaining the ProciGen rendering configuration.
GRAB supplies the posed body and hands together with the object trajectory, so
their relative 3D configuration is preserved, while ProciGen supplies clothed human
appearance and calibrated monocular views. This gives rendered images with
direct 2D-3D correspondence and articulated-hand ground truth. We subsample
PGG at 0.5\,fps to approximately 4K frame-views.
ARCTIC~\cite{Fan_2023_CVPR} is independently
subsampled at the same rate to approximately 5K frame-views. As illustrated in
Fig.~\ref{fig:pgg_examples}, the small image footprint of the people, hands,
and objects in PGG is intentional. PGG was designed to evaluate small hands and
objects, complementing ARCTIC, whose objects are relatively large and whose
hands are generally clearly visible.

\subsection{Dataset-specific comparisons}
Each baseline is evaluated where its representation and established protocol
support a meaningful comparison. For joint HOI
reconstruction, HDM~\cite{xie2023template_free} and
TeHOR~\cite{Nam_2026_CVPR} are evaluated with GraspHOI on ARCTIC and PGG.

TeHOR produces finite outputs for 85.5\% of the evaluation inputs. Most failed
cases occur when its detector finds no unique person or when a very small object
crop cannot be voxelized by TRELLIS~\cite{xiang2025structured}. A small remainder comes from missing inputs
or resource errors. The HOI table therefore averages every method over the
4{,}373 ARCTIC and 3{,}710 PGG frames on which all three methods return finite
predictions and a valid H+O score, and reports FL, PD, Offset, and PJO over
the hands that have contact and a valid TeHOR TRELLIS mesh.

Video-based HOI methods generally assume temporal consistency and use
multi-frame evidence unavailable to single-shot methods. Supplying full clips
would therefore give them additional input, while restricting them to one frame
would depart from their intended protocol. Neither yields a controlled
comparison. CARI4D~\cite{xie2025cari4d} presents further incompatibilities. It
assumes that the object is fully visible in the first frame, from which
Hunyuan3D~\cite{hunyuan3d2_2025} reconstructs its reference geometry, whereas
objects in our in-the-wild images are predominantly occluded. ARCTIC is also unsuitable
because its objects are articulated while CARI4D assumes rigid objects, and
PGG/GRAB objects frequently fall below CARI4D's object-size threshold, which is
designed for larger objects. We therefore exclude video-based methods from the
quantitative comparison.

For optimization-based reconstruction, PICO~\cite{Cseke2025PICO} and GraspHOI
are evaluated on PICO-db using FL, PD, and Offset. PICO requires contact
maps, which PICO-db provides. PICO-db does not provide posed 3D hand joints, so
we do not report PA-MPJPE, PJO, or 2D MPJPE for this comparison.
EasyHOI~\cite{Liu_2025_CVPR} and
GraspHOI are evaluated on ARCTIC and PGG using hand PA-MPJPE, FL, PD, Offset,
PJO, and 2D MPJPE.

\subsection{Optimization-baseline protocols}
EasyHOI reconstructs a single hand per image. Its rows and the corresponding
GraspHOI rows are therefore evaluated on the same recovered frames and
contacting hands: 3{,}701 on ARCTIC and 2{,}254 on PGG. These are subsets of
the approximately 5K ARCTIC and 4K PGG samples, retaining only frames in which
both methods recover the annotated contacting hand. Every metric in these rows is
computed on this hand: PA-MPJPE and 2D MPJPE on all of these hands, and FL, PD,
Offset, and PJO on those in selected frames that have contact.
We report Procrustes-aligned hand MPJPE for these rows because EasyHOI estimates
an independent camera for each image rather than using the dataset extrinsics.
Its predicted hand may therefore differ from the dataset frame by an arbitrary
rotation and scale even after wrist translation is removed. Per-hand similarity
alignment isolates articulated pose from this coordinate-system ambiguity. On
PICO-db, FL, PD, and Offset are paired over the 1{,}223 images with an
annotated contacting hand reconstructed by both methods. Some large objects such as chair, sofa, bed are excluded as they have hand contact but are mostly body-object contact. PA-MPJPE, PJO, and 2D MPJPE are omitted because the dataset provides no posed 3D hand joints.

EasyHOI's hand and object initialization can remain plausible on full-body
images, but their small image regions provide weak evidence for its subsequent
camera and object optimization. This is most pronounced on PGG, where EasyHOI's
estimated field of view reaches a median of $150^\circ$.

\subsection{Implementation and fairness}
We use released checkpoints and official inference settings without
test-set fine-tuning. Methods requiring a visible-object mask receive the same
mask, and methods requiring an object name receive the same dataset label.
Any method-specific retrieval, generation, or test-time optimization remains
part of that method's official pipeline. All outputs are scored using our
common implementation rather than values copied from the original papers.

We additionally note the configuration of TeHOR~\cite{Nam_2026_CVPR}. Its official repository states ``For the official paper setup, align the object mesh to the estimated depth map from ZoeDepth using ICP (Iterative Closest Point).'' and describes this step as optional. As the repository provides no implementation of this step, we run the public code without it. TeHOR's large object-placement errors in Table~\ref*{tab:hoi_baselines} of the main paper may therefore partly reflect this configuration.

For EasyHOI~\cite{Liu_2025_CVPR}, the method has an internal cropping mechanism included in its pipeline. As such, we feed it the same full frames as ours, without an extra input crop; EasyHOI is not capable of executing proper optimization on hands and objects that occupy too few pixels.

\subsection{Runtime}
We measure per-image runtime (Table~\ref{tab:supp_runtime}) with each method occupying a dedicated NVIDIA RTX
4090. GraspHOI takes 108\,s per image without the FoundationPose fallback and
122\,s with it, 113\,s on average over PICO-db, where the fallback runs on
34.4\% of the images. Two-hand cases take longer
because they require an additional hand-optimization step. PICO varies
substantially with object and mesh complexity, so we report its observed range
rather than a potentially misleading mean.

\begin{table}[t]
\centering
\small
\setlength{\tabcolsep}{10pt}
\begin{tabular}{lc}
\toprule
Method & Time per sample \\
\midrule
GraspHOI (Ours) & 113\,s \\
EasyHOI & 118\,s \\
TeHOR & 170\,s \\
PICO & 300--500\,s \\
HDM & 120\,s \\
\bottomrule
\end{tabular}
\caption{Per-sample runtime on a dedicated RTX 4090. All single values are
means. PICO is reported as a range.}
\label{tab:supp_runtime}
\end{table}

\subsection{Surface and alignment metrics}
We uniformly sample 10K points from each available mesh surface. A predicted
point cloud, such as HDM's object, is sampled directly. Shape Chamfer distance
(sCD) is the average of the two directed nearest-neighbor distances. For
the HOI comparison, we center each prediction and ground-truth scene at its own
pelvis. The same translation is applied to the human and the object, so relative
human-object placement remains in the metric. The sCD of the human, of
the object, and of their combined samples (H+O) is computed in this frame and reported in mm. No part-specific
translation, rotation, or scale alignment is applied, so sCD reflects shape, scale, orientation, and placement relative to
the person together. H+O pools the human and object samples, so a misplaced object can still match ground-truth human points; object sCD therefore measures object placement more directly.

Body
MPJPE is computed over the 22 SMPL body joints after pelvis centering. In the joint
metrics of the HOI and optimization-baseline tables, every joint, predicted and
ground truth alike, is regressed from the posed mesh. Body joints use the SMPL-H regressor for the
predictions and the subject's gendered SMPL-X model for the ground truth. The 16
hand joints, \ie the wrist and 15 finger joints, are regressed with MANO's
regressor from the 778 hand vertices of every model, so that all hands share
MANO's wrist. In the HOI table, hand MPJPE is computed over the
wrist and 15 articulated finger joints of each annotated contacting hand in the
same pelvis-aligned frame, without wrist or Procrustes alignment, so it includes
where the arm places the hand. Both the HOI and the optimization-baseline tables report per-hand
PA-MPJPE: a similarity transform is fitted on the same 16 joints, and the error
is averaged over the 15 finger joints.

\subsection{HDM fitting and scale alignment}
HDM reconstructs an up-to-scale human-object point cloud. For human mesh and
joint metrics, we fit SMPL-H to its human points using the correspondence
autoencoder and fitting procedure released with InterTrack~\cite{xie2024intertrack}.
We then resolve the one
global gauge per frame from the common 22-joint SMPL body skeleton:
\begin{equation}
s_i =
\frac{\sum_{(j,k)\in\mathcal B}\lVert J^{\mathrm{GT}}_{i,j}-J^{\mathrm{GT}}_{i,k}\rVert_2}
     {\sum_{(j,k)\in\mathcal B}\lVert J^{\mathrm{HDM}}_{i,j}-J^{\mathrm{HDM}}_{i,k}\rVert_2},
\end{equation}
where $\mathcal B$ contains the 21 SMPL body bones. Summed bone length is
pose-invariant, unlike a posed bounding-box height. We center the HDM scene at
its predicted pelvis and multiply both the human and object by the same $s_i$.
The ground-truth scene is centered at its own pelvis. We apply no rotation,
ICP, or object-specific scale/translation alignment. Thus, HDM retains its
predicted relative human-object scale and placement. FL, PD, Offset, and PJO need an inside-outside test, so for
these metrics the scaled object points are meshed by screened Poisson
reconstruction~\cite{kazhdan2013screened} with Open3D's default settings, \ie
octree depth 8 and normals from a plane fit to the 30 nearest points, and
without density trimming. The normals are oriented away from the cloud centroid,
and the scaled SMPL-H fit supplies HDM's hands.

\section{Hand and Contact Metrics}
We provide additional details for the hand-object metrics used in our experiments.

\textbf{Float, Penetration Depth, and Offset.}
We evaluate only the hand or hands marked as contacting in the dataset
annotation, and score each contacting hand separately. Let $d(v)$ be the
winding-number signed distance from a vertex $v$ of the 778-vertex hand region
to the object, positive outside. If any vertex penetrates ($d(v)<0$), penetration
depth (PD) is the mean of $-d(v)$ over the penetrating vertices and float (FL) is
zero. Otherwise, FL is the hand-object gap $\min_v d(v)$ and PD is zero. Offset
is FL$+$PD. All three are reported in mm and averaged over every contacting
hand, so floating hands contribute zero PD and penetrating hands contribute zero
FL. The original hand vertices are used for FL and PD. Contact (\%) in Tables~\ref*{tab:stagewise} and~\ref*{tab:joint_ablation} of the main paper is the
percentage of hands whose Offset is at most 3\,mm, the distance within which
ARCTIC~\cite{Fan_2023_CVPR} counts a hand vertex as in contact. A floating hand
thus counts if its gap is at most 3\,mm, and a penetrating hand if the mean
depth of its penetrating vertices is at most 3\,mm.

For SMPL-H, we select hand vertices using a semantic rule: cumulative
skinning weight greater than 0.5 for the wrist and 15 articulated hand joints.
This produces a 778-vertex region per side. For SMPL-X, we use the 778 vertices
per side of the official MANO-to-SMPL-X vertex correspondence. No SMPL-H-to-SMPL-X surface
conversion is applied during scoring.

\textbf{Per-Joint Offset (PJO).}
For each hand, PJO uses 21 points: the 16 MANO joints (wrist and 15 finger
joints), regressed from the 778 hand vertices with MANO's joint regressor, and
the five MANO fingertip vertices. In the prediction and in the ground truth
separately, each point receives its signed distance $d$ to that scene's own
object, as defined above. PJO is the mean of $|d_{\mathrm{pred}}-d_{\mathrm{GT}}|$
over the 21 points, in mm, so a point 3\,mm inside the object where the ground
truth has it 3\,mm outside gives a difference of 6\,mm. Each scene is measured in its own
frame, with no alignment other than the HDM scale alignment described above.
PJO is averaged over the same hands as FL, PD, and Offset.

\textbf{2D MPJPE.}
In the optimization-baseline comparison, the 16 MANO joints of the evaluated
hand, regressed as for PJO, are projected with each method's own camera, \ie
the intrinsics $K$ for GraspHOI and EasyHOI's estimated camera, and the
corresponding ground-truth joints with the dataset camera. 2D MPJPE is the mean
distance between the two projections, in pixels at the native image resolution:
$2000\times2800$ or $2800\times2000$ on ARCTIC and $2048\times1536$ on PGG. It
uses the same hands as PA-MPJPE.

\section{Extended Qualitative Results}

\begin{figure*}[p]
  \centering
  \includegraphics[width=0.9\textwidth,height=0.86\textheight,keepaspectratio]{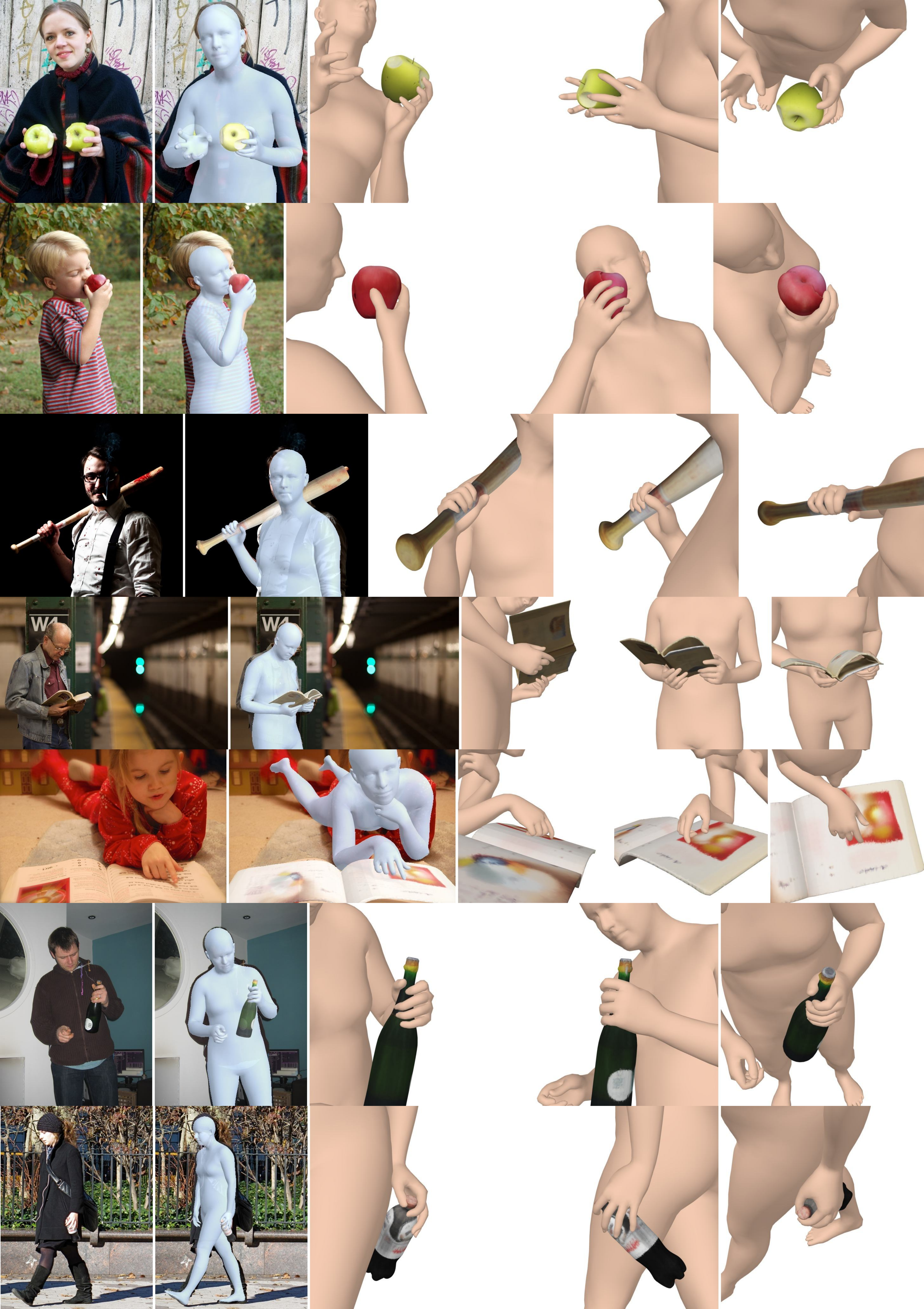}
  \par\smallskip\textbf{(a)}
  \caption{Extended qualitative results on PICO-db. Across panels (a)--(c), each row shows the input, the reconstruction overlaid on it, and three alternate 3D views. These examples are qualitative and include large objects that the quantitative PICO-db evaluation excludes.}
  \label{fig:qual_supp}
\end{figure*}

Figure~\ref{fig:qual_supp} extends the
qualitative evaluation to additional interactions.

\section{Failure Cases, Limitations, and Future Work}
\label{sec:limitations}
We present representative failures caused by upstream initialization and by the
subsequent optimization stages. Because the pipeline is decoupled, an error in
one reconstructed component can remain geometrically plausible in the image
while becoming incompatible with contact in 3D.

\begin{figure}[!tbp]
\centering
\includegraphics[width=\linewidth]{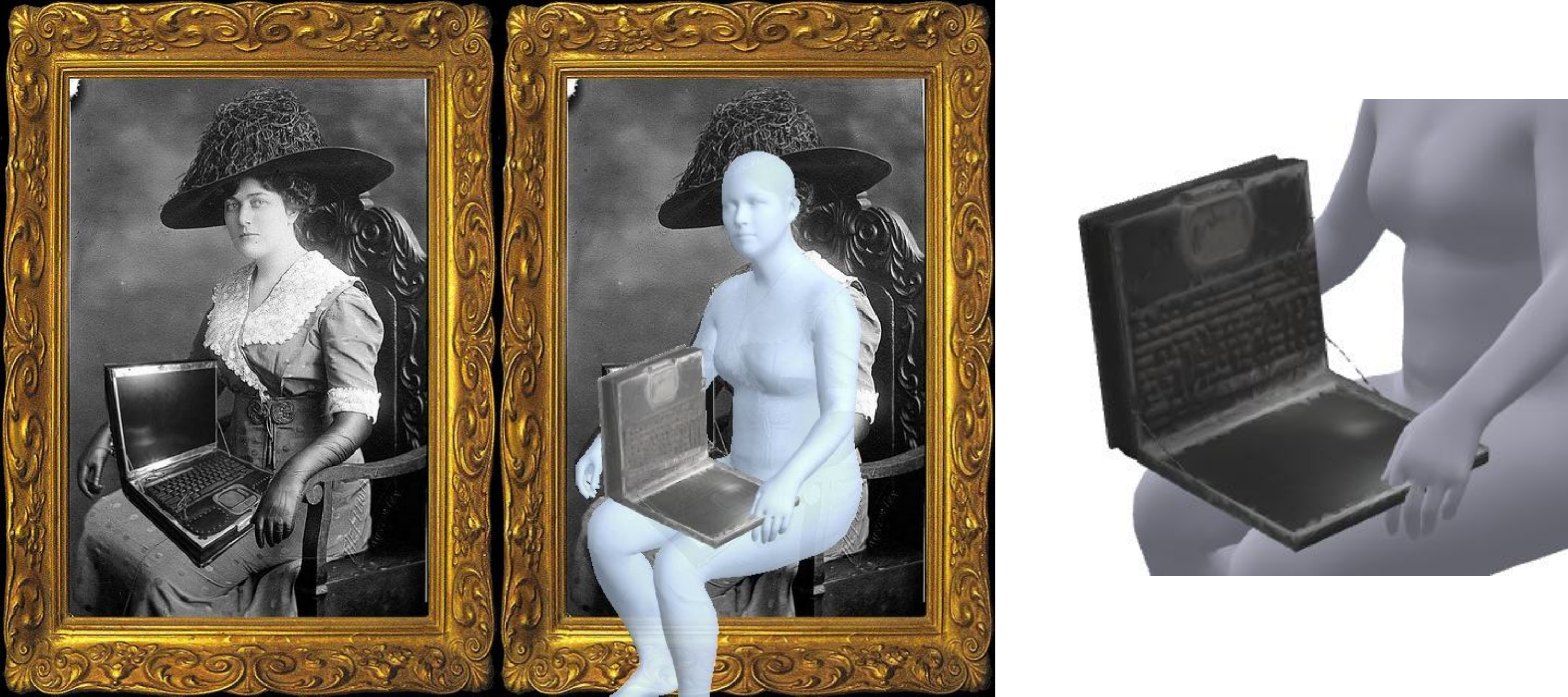}
\caption{Fallback-registration failure. FoundationPose can return an
incorrect pose when the mesh differs from the image evidence or its
initial metric scale is inaccurate.}
\label{fig:object_pose_failure}

\vspace{\floatsep}
\includegraphics[width=\linewidth]{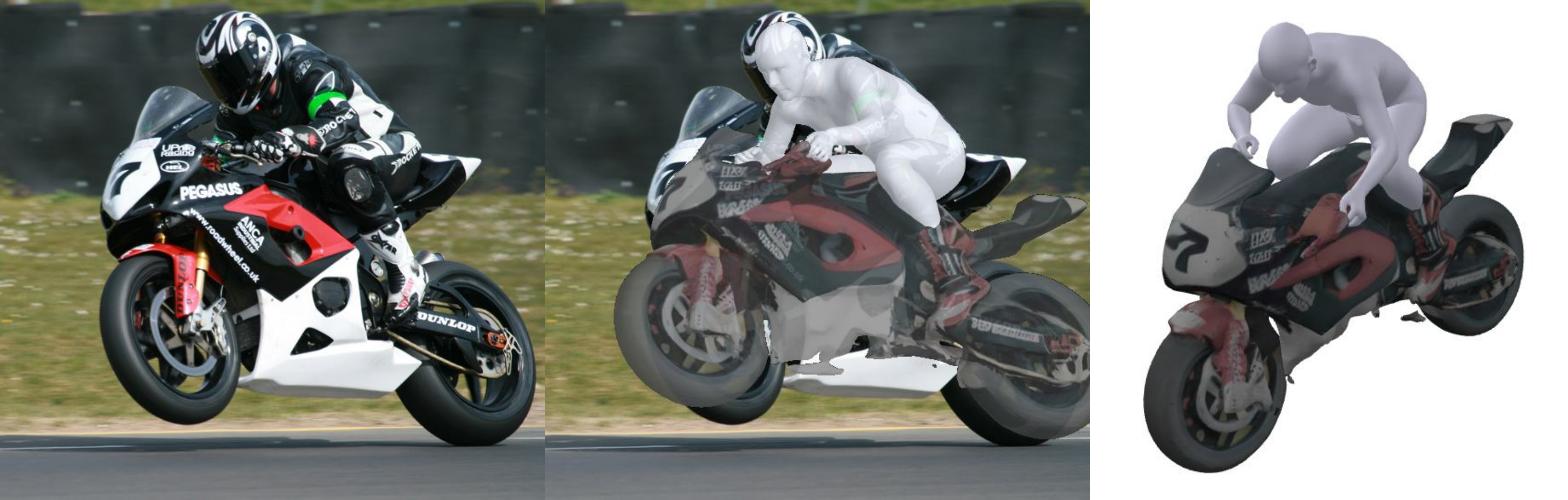}
\caption{Joint-optimization failure. Because refinement freezes the body
outside the optimized arm-hand chains, a large body-contacting object can
penetrate the body.}
\label{fig:joint_body_failure}

\vspace{\floatsep}
\includegraphics[width=\linewidth]{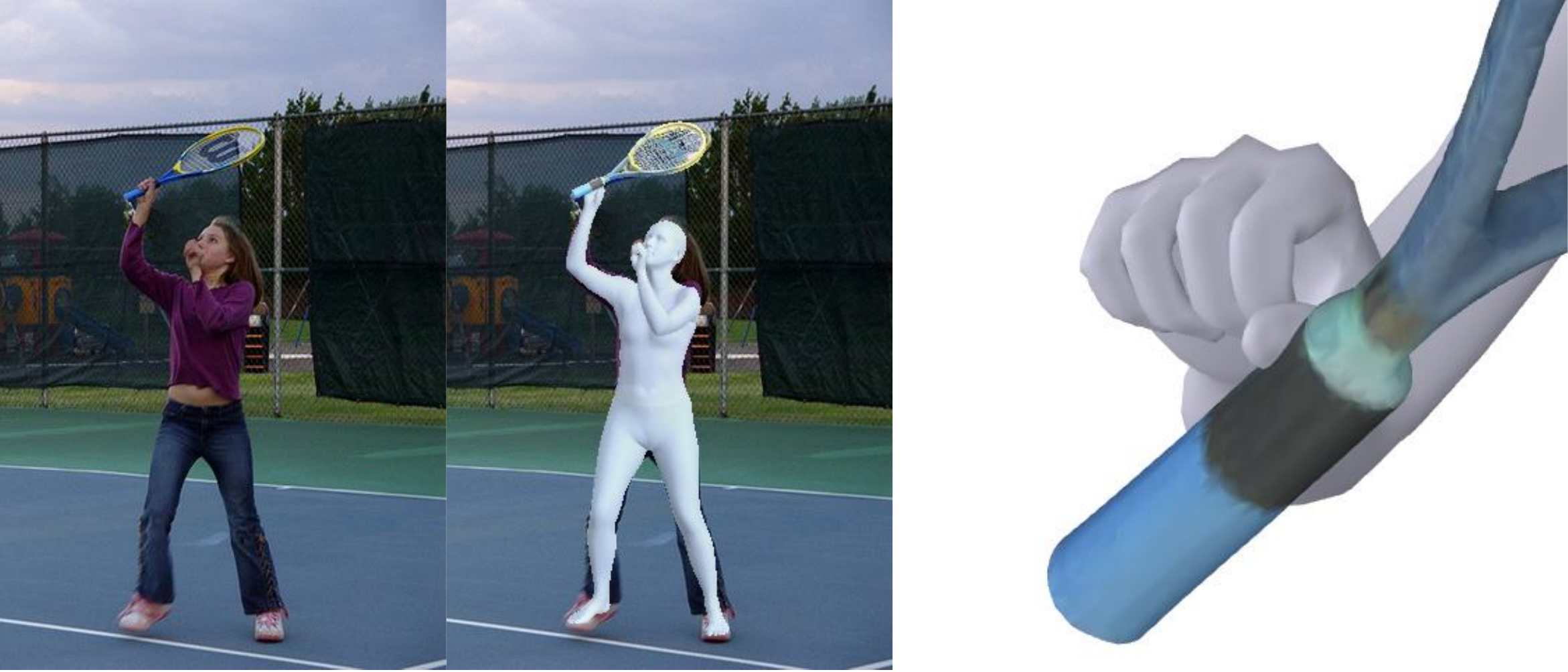}
\caption{Grasp-opening failure. The racket handle is thicker than the opening
of the initialized grasp, so it never enters the hand and stays outside.}
\label{fig:grasp_opening_failure}
\end{figure}

\subsection{Upstream Failures}

As our method relies on the performance of foundation models, one of the main bottlenecks is errors from the models themselves. SAM 3D Objects~\cite{sam3dteam2025sam3d3dfyimages} does not always recover the observed geometry. The placement steps correct the position, orientation, and
size of the mesh but not its shape, so a mesh that is too thick or wrongly
proportioned can match $S_a$ in projection and still be incompatible with a
collision-free grasp. Every placement step also treats $S_a$ as its target. When
the amodal inpainting hallucinates object parts or misses occluded ones, the
re-fit and the silhouette refinement fit the wrong silhouette.

\subsection{Fallback Registration Failures} \label{sec:object_fail}

One of the main issues arises from incorrect object pose initialization. Layout tokens from SAM 3D Objects~\cite{sam3dteam2025sam3d3dfyimages} often predict flipped poses for semi-symmetric objects. This cannot be filtered with our IoU gate due to the nature of the object. Likewise, our fallback FoundationPose~\cite{wen2024foundationpose} is prone to two problems. First, the mesh may not match the object in the input closely enough for reliable registration. Second, FoundationPose is sensitive to object-scale initialization, while our monocular scale estimate does not always
recover the correct metric size. Either mismatch can lead to an incorrect object
pose, as shown in Fig.~\ref{fig:object_pose_failure}. 

\subsection{Joint Optimization Failures}

Joint refinement freezes the body outside the optimized arm-hand chains while
placing greater emphasis on hand alignment. This is effective for grasp-centric
interactions but cannot correct the body around a large body-contacting object.
Such objects may therefore penetrate the fixed body even when the hands are
plausibly aligned, as shown in Fig.~\ref{fig:joint_body_failure}.

Joint refinement also cannot widen the initialized grasp. Finger joints may
only turn a few degrees from their initial rotations
(Sec.~\ref{sec:supp_joint}), so the opening between the fingers and the palm
stays close to that of the initialized hand. When the object is thicker than
this opening, it never makes it into the hand, because the penetration terms keep it
from passing through the fingers. As a result, it stays outside the grasp, as shown in
Fig.~\ref{fig:grasp_opening_failure}.

The contact correspondences also assume that the palmar side of the hand touches the object wherever the two overlap in the image. In fingertip grasps, this can pull the palm toward a surface that it does not touch.

\subsection{Current Scope}
GraspHOI targets static, grasp-centric interactions with at least one contacting
hand. We target a more specific goal, and we expect body-object contact to be inferred from upstream tasks. Our method does not model temporal motion, deformable-object physics, or
large-scale body-object support such as sitting and leaning. It also requires a
visible-object mask and a noun prompt for amodal inpainting. Errors in SAM 3D
Body~\cite{Yang_2026_CVPR}, WiLoR~\cite{potamias2024wilor}, depth estimation,
amodal conditioning, object reconstruction, or contact detection can propagate
into the final scene. In particular, monocular
symmetry and scale-depth ambiguity may survive image-space alignment, while an
incorrect generated thickness can make the image evidence incompatible with a
collision-free grasp. Occlusion is a further limitation, as we have no temporal information to guess the occluded hands or object. When the object is largely hidden, most of $S_a$ must be generated rather than observed, and every placement step fits this partly generated
silhouette (Sec.~\ref{sec:supp_obj}). When the grasping hand is largely hidden,
WiLoR can misread it, and a hand that fails the direction check of
Sec.~\ref{sec:supp_hands} has no trusted keypoints and is excluded from joint
refinement. Finally, the manually weighted refinement objectives may
require recalibration under a substantial domain shift.

Additionally, while our method reconstructs realistic grasps better than previous methods, we believe that grasps are not yet fit to be used in the field of robotics, unlike hand-object interaction works that focus mainly on the fingers without awareness of the arm poses. Sim2real is not applicable to our method, and as such we do not perform any physics evaluation. We expect our work to be a step toward full-body 4D HOI reconstruction pipelines.

\subsection{Future Work}
These limitations suggest several directions. Temporal evidence from video
could fill in occluded object parts and resolve the remaining scale-depth
ambiguity, and the reconstructed static interactions can serve as keyframes for
HOI video generation. Jointly refining the body with the arm-hand chains would
handle large body-contacting objects, and modeling deformable objects and
body-object support such as sitting and leaning would widen the range of
interactions. Finally, learning the refinement weights or a contact prior from
data could replace the manual weighting and its recalibration under domain
shift.

\FloatBarrier

\section{Licensing Details}
\label{sec:license}

\begin{itemize}
    \item \textit{``Basketball portrait \#2"} by rady one is licensed under CC BY~2.0.
    \item \textit{``Mariachi singer, Chicago, Illinois, USA - 20060814"} by Codo is licensed under CC BY-SA~2.0.
    \item \textit{``the pineapple"} by sunshinecity is licensed under CC BY~2.0.
    \item \textit{``Tim Beavers lead guitar \& vocalist for PBR (People's Blues of Richmond) \#rva"} by Sky Noir is licensed under CC BY-NC~2.0.
    \item \textit{``GIRL ft.\ ICECREAM"} by Hsuanya Tsai is licensed under CC BY-NC~2.0.
\end{itemize}

Dataset examples are drawn from
ProciGen-GRAB~\cite{xie2023template_free,Taheri2020GRAB},
PICO-db~\cite{Cseke2025PICO}, ARCTIC~\cite{Fan_2023_CVPR}, and
InterCap~\cite{huang2024intercap}. Internet images used in
the retained figures are listed above.

\clearpage
\clearpage
\begin{figure*}[p]
  \centering
  \includegraphics[width=0.9\textwidth,height=0.9\textheight,keepaspectratio]{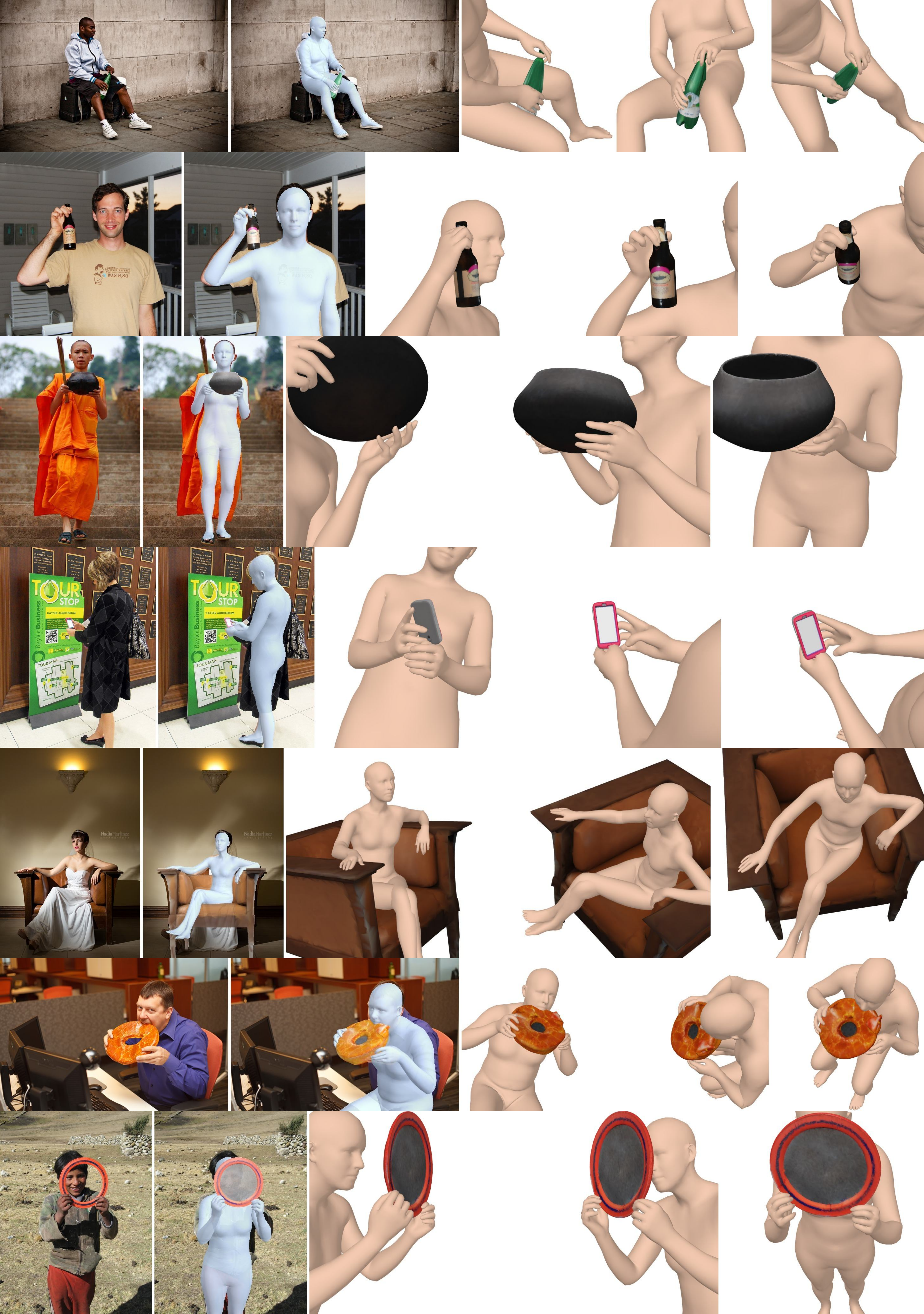}
  \par\smallskip\textbf{(b)}
\end{figure*}
\clearpage
\begin{figure*}[p]
  \centering
  \includegraphics[width=0.9\textwidth,height=0.9\textheight,keepaspectratio]{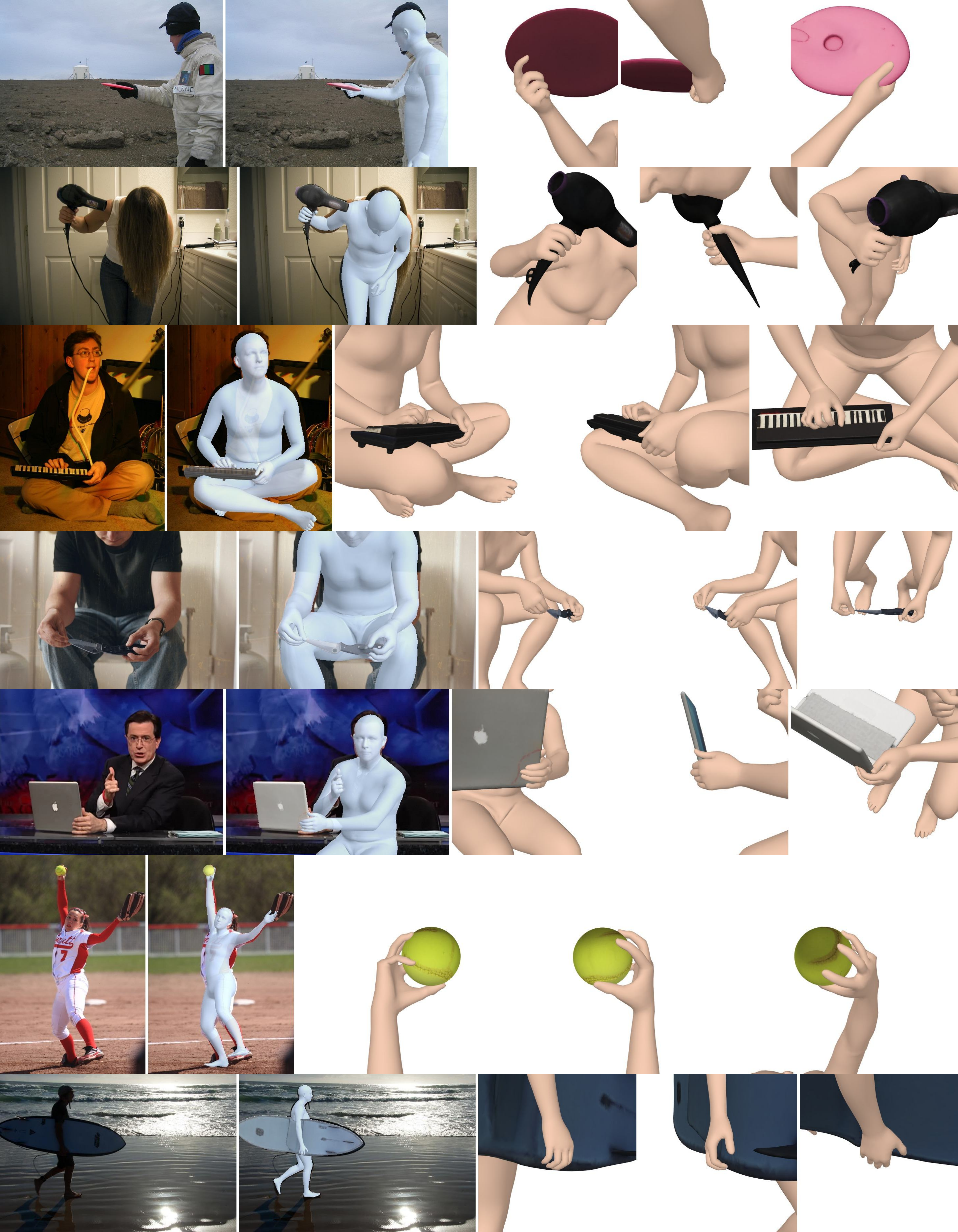}
  \par\smallskip\textbf{(c)}
\end{figure*}

\clearpage
\clearpage
\begin{figure*}[p]
  \centering
  \includegraphics[width=0.9\textwidth,height=0.9\textheight,keepaspectratio]{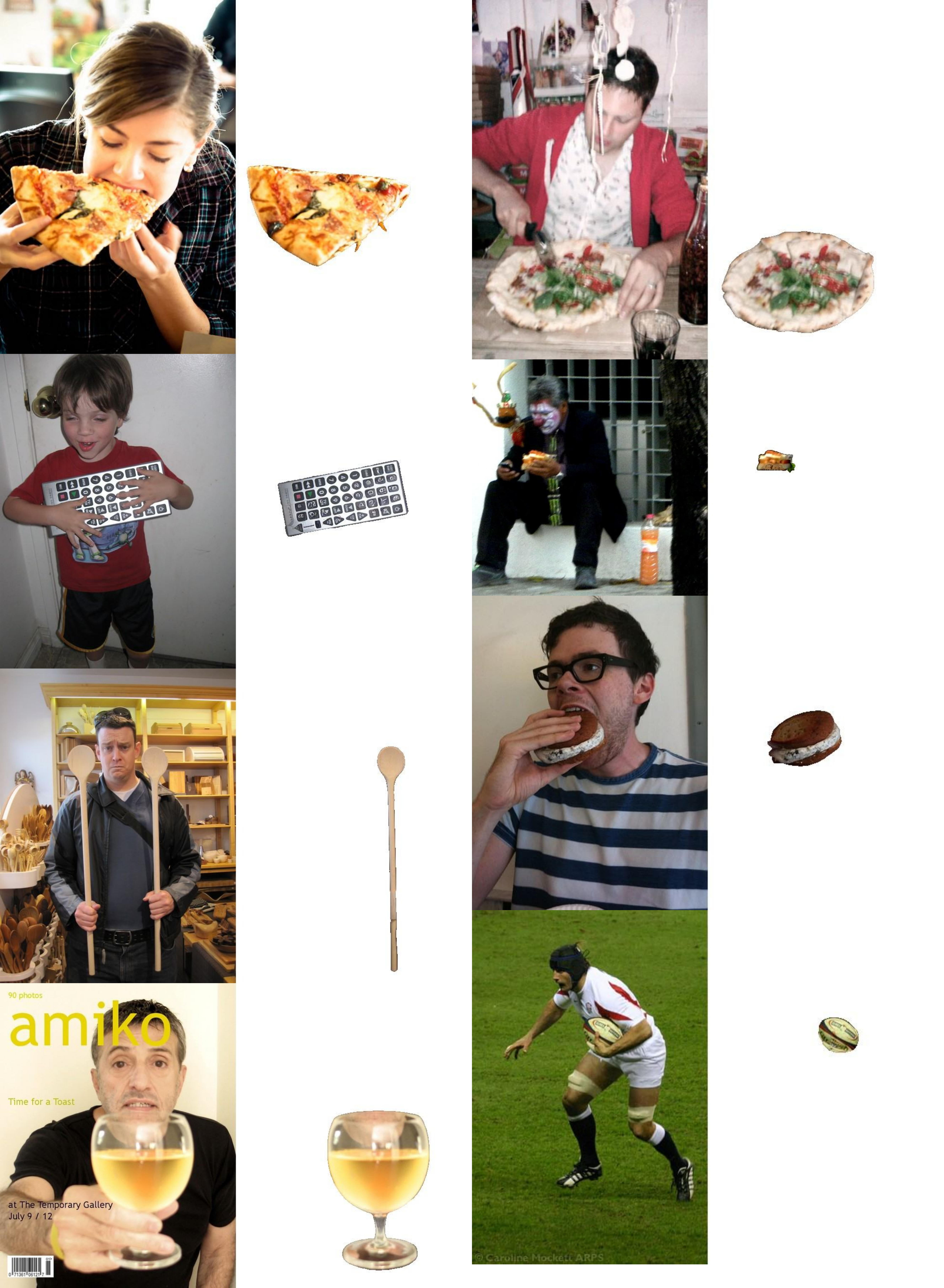}
  \par\smallskip\textbf{(b)}
\end{figure*}
\clearpage
\begin{figure*}[p]
  \centering
  \includegraphics[width=0.9\textwidth,height=0.9\textheight,keepaspectratio]{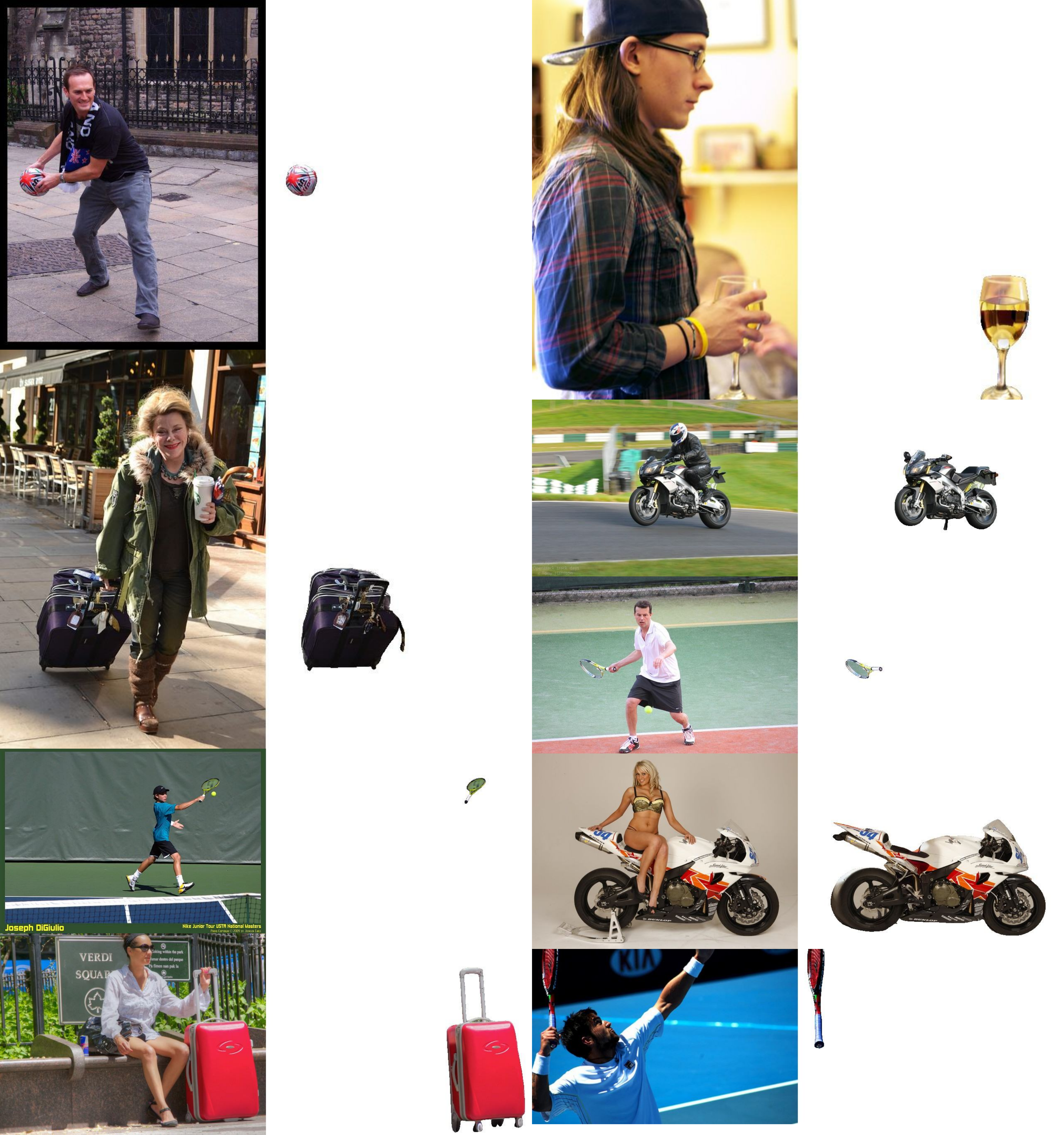}
  \par\smallskip\textbf{(c)}
\end{figure*}
\clearpage

\end{document}